\documentclass{article}
\usepackage[utf8]{inputenc}
\usepackage{amsmath}
\usepackage{amsfonts}
\usepackage{amsthm}
\usepackage{graphicx}
\usepackage[colorinlistoftodos]{todonotes}
\usepackage[colorlinks=true, allcolors=blue]{hyperref}
\usepackage{algorithm}
\usepackage[noend]{algpseudocode}
\usepackage[normalem]{ulem}
\usepackage{bbm}
\usepackage{caption}
\usepackage{subcaption}
\usepackage{cancel}
\usepackage{fancyhdr}
\usepackage[margin=2.25cm]{geometry}

\newcommand{\Delete}[1]{}
\graphicspath{ {./images/} }
\newcommand{\bs}[1]{\boldsymbol{#1}}

\title{Machine Learning based Framework for Robust Price-Sensitivity Estimation with Application to Airline Pricing}

\author{Ravi Kumar\footnote{corresponding author} \footnote{authors contributed equally}, Shahin Boluki\footnotemark[2], Karl Isler, Jonas Rauch, Darius Walczak}

\date{
   PROS Inc., Suite 600, 3200 Kirby Dr, Houston TX 77098 USA\\%
    \today
    }

\begin{document}
\maketitle 

\abstract{\textbf{\textit{Problem definition:}} We consider the problem of dynamic pricing of a product in the presence of feature-dependent price sensitivity. Developing practical algorithms that can estimate price elasticities robustly, especially when information about no purchases (losses) is not available, to drive such automated pricing systems is a challenge faced by many industries. \textbf{\textit{Methodology:}} Based on the Poisson semi-parametric approach, we construct a flexible yet interpretable demand model where the price related part is parametric while the remaining (nuisance) part of the model is non-parametric and can be modeled via sophisticated machine learning (ML) techniques. The estimation of price-sensitivity parameters of this model via direct one-stage regression techniques may lead to biased estimates due to regularization. To address this concern, we propose a two-stage estimation methodology which makes the estimation of the price-sensitivity parameters robust to biases in the estimators of the nuisance parameters of the model. In the first-stage we construct estimators of observed purchases and prices given the feature vector using sophisticated ML estimators such as deep neural networks. Utilizing the estimators from the first-stage, in the second-stage we leverage a Bayesian dynamic generalized linear model to estimate the price-sensitivity parameters. \textbf{\textit{Results:}} We test the performance of the proposed estimation schemes on simulated and real sales transaction data from the Airline industry. Our numerical studies demonstrate that our proposed two-stage approach reduces the estimation error in price-sensitivity parameters from 25\% to 4\% in realistic simulation settings. \textbf{\textit{Managerial implications}}: The two-stage estimation techniques proposed in this work allows practitioners to leverage modern ML techniques to robustly estimate price-sensitivities while still maintaining interpretability and allowing ease of validation of its various constituent parts.}


\section{Introduction}\label{sec:introduction}

\subsection{Background and Overview}
Many sellers are interested in improving their pricing decisions by dynamically adjusting prices of their products based on unique product features and other relevant information available at the time of request e.g., competing product prices, market indices etc. Automated dynamic pricing systems that can enable such functionality typically require estimates of customers' price-sensitivity at this granular level using historical sales transaction data. Although modern IT systems have made it possible to collect and use rich request-specific information from a shopping session, the estimation of price-sensitivity from such feature-rich historical observational data still remains a challenging task in many industries \cite{hua2021kdd, ye2018kdd}. This is especially true when the firm is only able to observe the total quantity sold for a product at a certain price in given time period but is unable to observe customer requests that did not result in a sale. Data about unsuccessful sale events at an individual request level is commonly referred to as loss or no-purchase information and if available can be utilized to estimate price-sensitivity by directly modeling purchase probabilities as a function of price and other relevant features via methods like parameterized logistic regression. 
Situations where loss information is not available are common in industries like Airlines, Car Rentals, and Hotels where a large fraction of sales occur via third-party websites or Online Travel Agents (OTA) such as Expedia, KAYAK, Google, Orbitz etc. Typically, a firm selling its products via these OTAs does not have direct visibility into loss information corresponding to a large fraction of individual requests on these platforms. In the absence of such loss information, one may estimate price-sensitivity by learning price-demand relationship (demand response modeling) using total sales for a product in given time period e.g., daily sales.

In this work we focus on learning such price-demand relationship given historical observations. This is akin to an observational study setup where the purchases are observed under the historical pricing policy deployed by the firm. A na\"ive approach can be to fit a supervised model with desired features (including price-related) and demand as the response variable. But this approach has several shortcomings. 
One of the shortcomings is related to the choice of the model. On one hand, parametric models that are popular in the literature provide an explainable and easy to use model with interpretability of their parameters, but are limited in terms of their flexibility and have strong assumptions about the generating process. On the other hand, modern machine learning (ML) approaches such as neural networks and gradient boosting, despite their high predictive power, do not easily lend themselves to constructing an interpretable framework for the inference task related to the price sensitivity estimation. 

Another issue of directly fitting a supervised model is the presence of confounders, which are features that affect both demand and price. It is well known that omission of such features from the regression model may lead to biased estimates of price sensitivity parameters when estimated from observational data. In addition, although our main goal in this paper is not demand prediction, it is worth mentioning that a complex predictive model directly fitted on demand using price and additional features cannot generally make reasonable demand predictions for prices outside the range of observed values or prices different from the pricing policy underlying the observed data, as those regions are out of the training data distribution \cite{zou2020counterfactual,shalit2017estimating}. Explainable and simpler parametric models based on micro-economic theory and domain specific knowledge may be less susceptible to this challenge compared with more complex approaches such as deep neural networks. 

Our approach for demand response modeling is thus based on a hybrid framework where we utilize modern ML-based non-parametric techniques such as deep neural networks to make an observational baseline prediction of both price and demand, and then combine it with a parametric approach to estimate the causal relationship between deviations of price and demand from this baseline. To this end, we build upon recent advances in the area of causal inference and semi-parametric regression in econometrics \cite{Athey2019} and employ related ideas to construct robust price-sensitivity estimators using Poisson semi-parametric modeling. Moreover, by utilizing variational Bayesian approach for updating the model parameters in an online fashion, this approach is compute and storage efficient from a practical implementation perspective.

\subsection{Pricing in Airline Industry and Challenges}

In the airline dynamic pricing application motivating this work, a product being priced is typically an itinerary defined by specific route (Origin, Destination and Path), trip type (round trip, one way, multi-city etc), date and time of departure. Moreover, airlines typically differentiate products by compartment (economy, economy-premier, business, first etc.) and associated privileges and restriction e.g., change-related fees, upgrade eligibility, baggage allowance etc. Additional trip-related features are also used to differentiate customer price-sensitivity, these may include duration of stay at the destination, advance purchase restrictions, departure day of week, and number of layovers. 

Airline demand is significantly impacted by yearly seasonal patterns; for example often there is more demand in summer months due to school holidays. Moreover, special events (HSE) such as conferences in the origin or destination city also have a major impact on demand. Demand patterns also vary over the booking horizon of an itinerary. Most airlines allow the booking process to start around one year before the departure date. The pattern of demand over this booking horizon can show significant variation due to the fact that leisure customers, who represent a large fraction of demand, are highly price-sensitive, and book their trips early while business travelers with higher willingness-to-pay (WTP) may book their trip much later. Therefore, causal inference of price-sensitivity parameters via price-demand relationships requires inclusion of features to control for yearly seasonality, booking days prior and HSEs in the model.

The competitive landscape that the airline is operating in also plays an important role in determining its demand. On most routes there are multiple airlines offering services. The set of competing products that may impact an itinerary's demand are not only the similar (and relevant) itineraries offered by a competing airline but also other products for the same origin-destination pair offered by airline itself. Moreover, quality indicators such as  on-time performance or rating of in-flight services could also be correlated to price-sensitivity.  

Unlike many scenarios related to e-commerce setting where the inventory costs for the product being sold are minimal due to supply being relatively elastic, airlines have to deal with fixed supply of seat inventory on flights in their network. Any seat not sold by the departure date brings no revenue and gets wasted i.e., the inventory is perishable. To deal with the fixed and perishable inventory, airlines employ sophisticated Revenue Management (RM) systems. Prices generated by an airline for an itinerary are the product of complex interactions between the RM system output and further refinements done via pricing rules, also known as fare rules in the airline industry. Airline RM systems work at a higher (aggregated) level than the level at which an itinerary is priced. For example, regardless of whether an RM system performs optimization at the leg or the origin-destination (OD) level, round-trip related dimensions e.g., duration of stay, are not explicitly considered in the RM system optimization process. RM systems determine the expected marginal opportunity costs involved in fulfilling the itinerary request by  considering product prices at an aggregated level (fare classes), forecasts of future demand, remaining seat inventories and time until departure. These marginal opportunity costs, also known as bid prices in the airline industry, are used to determine availability of fare classes at the OD level. Further refinement to get to the level of itinerary price is handled via a complex set of fare rules tied to these fare classes \cite{Vinod2021}. Often these fare rules are not fully data-driven based on an empirical understanding of customers' price-sensitivities, but rely to a large extent on experience of analysts and managers in the airline's pricing departments. 

In this work, we focus on developing more data-driven approaches to refine prices given the aggregate level marginal opportunity costs (bid prices) computed by the RM system. In particular, we assume that the firm has an RM system to perform finite-horizon network level optimization at an aggregate level and therefore it can compute the marginal opportunity costs, $c_{it}$, associated with fulfilling an itinerary request $i$ that it receives at time $t$. Given this cost and an $m$-dimensional feature-vector, $\mathbf{X}_{it}$, associated with this itinerary request, the firm is interested in further refining the price to charge for this request. Features associated with the itinerary can include information about origin, destination, number of stops, point of sale, departure time, day of week, holiday indicators etc. 

To enable such dynamic pricing policies, we develop robust methods for estimating the price-sensitivity of customer population for such featurized itineraries via demand response modeling. Note that this system for further price refinement is separate from the RM system and only takes as input marginal opportunity costs generated by the RM system i.e., the problem of interest in this work is to find an optimal price for the itinerary given that the marginal opportunity cost has already been computed by the RM system. The RM and pricing systems are implicitly connected as the bookings generated and prices associated with sales transactions are fed back into the RM system forecaster at an aggregate level to forecast class level demands and fares used for network optimization. We assume that the firm takes into consideration some consistency requirements between the two systems while generating inputs for the RM system e.g., by appropriately mapping booking and price transactions to aggregated level products (classes) used in the RM system, pro-rating fares from round-trip to half-return level etc. This interaction between the RM and Pricing systems although important are beyond the scope of this work and here we solely focus on the price sensitivity estimation problem.

Given the complex RM system and pricing methods used by the airline, the price offered for an itinerary is correlated with many of the features we have discussed earlier in this section. What makes estimating price-sensitivity parameters via demand response modeling challenging is that the demand for an itinerary is also correlated with many of the same features that impact its price. For example, higher demand in summer months leads to higher prices due to the opportunity costs (or bid prices) for that time period being higher. Such features that affect both the price (treatment variable) and sales (outcome variable) are known in econometrics literature as confounders. If the confounders are not identified and included in the model i.e., they are not controlled for, a na\"ive estimation of price-sensitivity coefficients will be biased due to the so-called ``omitted variable bias'' \cite{Angrist2009}. This bias problem cannot be solved just by using more sophisticated ML based estimators. Instead, one would need to rely on Instrumental Variable (IV) techniques or conduct randomized pricing experiments by assigning randomly chosen prices to itinerary requests \cite{Angrist2009}.

Randomized pricing-based studies to estimate customers' price-sensitivity in the airline industry may also not be possible. The main reason is that such experiments could be perceived to be quite costly in the short-term because of using ``non-optimal'' pricing policy and fear of losing revenue. Besides, if two booking requests with the same features are offered randomly different prices, the passengers may try to get around the randomization by trying several times to get a lower price. Even worse, the randomization approach can be perceived as unfair and dishonest.

Given these challenges, industries like airlines have found it difficult to implement a robust framework for automated dynamic pricing. The current state of affairs has necessitated the use of complex rules-based pricing systems as the de facto solution to dynamic pricing in this industry. In our view, sophisticated ML techniques by themselves cannot solve this problem. However, combining these techniques with the recent advances in the area of causal inference allows for a more data-driven solution to this problem.

\subsection{Related Work}

This work is related to the literature on dynamic pricing in the presence of contextual features, and the literature on causal inference in the presence of treatment heterogeneity with applications to price elasticity estimation.

Constructing price-demand relationships for goods has a long history in economics, specifically in macro-economic theories related to the study of supply and demand. The use of demand response modeling in the more applied area of dynamic pricing via maximizing profit or revenue functions is relatively more recent but has already lead to a substantial amount of literature. The survey paper by \cite{denBoer2015} provides an extensive overview of the dynamic pricing and learning literature. The majority of the work in this area focuses on parametric models, where the functional form of the demand function is assumed to be known. This includes more recent works on dynamic pricing in the presence of contextual features \cite{Cohen2016, Qiang2016, Javanmard2019a, ban_personalized_2021,  Elmachtoub2021}. There is also some work in the area of non-parametric models which do not assume a known functional form for the demand model, see for example \cite{Besbes2009, Besbes2015} and \cite{Perakis2019}. However, most of these works have not focused on robust estimators for price-sensitivity from the causal inference perspective in the presence of many confounding variables, a situation common in industries like Airlines. There has also been work on semi-parametric models especially in the econometrics literature \cite{Robinson1988, Newey2004}. These models assume a parametric (often linear) structure for treatment effect and a non-parametric structure for the remaining part of the model. However, most of this line of work has focused on additive noise structure for the model of interest and results for demand models using discrete distributions like Poisson or Negative Binomial has been largely lacking. Moreover, there are new challenges related to regularization induced bias in the estimation of parameters of interest if the non-parametric part is modeled using modern ML techniques \cite{chernozhukov2018double}.

For estimating treatment effects e.g., price sensitivities, Instrumental Variable (IV) based two-stage estimation schemes are often used in the causal inference literature to reduce estimation bias \cite{pearl2009causality,newey2003instrumental,hartford2017deep,kmenta2010mostly}. An IV is a variable that directly affects the treatment (e.g., price) but not the outcome (e.g., bookings), i.e., its effect on the outcome is only through the treatment. IVs help in estimating the causal relationship from regression models in the presence of endogenous regressors, for example due to omitted confounders. Two-stage least squares (2SLS) estimation \cite{angrist1996identification} is one of the most widely known IV techniques which has a linear and homogeneous treatment effect assumption. In the first stage of 2SLS, the endogenous variable is regressed on the IVs to create a new latent variable, and in the second stage the endogenous variable is replaced by the new variable (prediction from the first stage model) in the main linear regression model of the outcome. Non-parametric IV frameworks have been developed for more general settings than the linear model \cite{newey2003instrumental,darolles2011nonparametric,chen2012estimation,hartford2017deep}. In \cite{hartford2017deep}, the authors train deep neural networks to model the distribution of treatment given the IVs and other features in the first stage and the outcome prediction model in the second stage. Although IV based frameworks are promising, identifying IVs itself is a challenging problem which limits their application.

With firms having access to richer data, it is possible to include most of the important confounders into the regression model. However, this results in a  high-dimensional setting and may also require a more sophisticated regression model. For such settings, Double or Orthogonal machine learning (DML) is another line of causal inference methods that use a two-stage estimation process \cite{chernozhukov2018double}. This approach assumes a partially linear model where the treatment effect is linear but other features are either high-dimensional (with respect to number of samples) and/or their effect on treatment and outcome is non-parametric. It combines sophisticated machine learning techniques for constructing estimators for the non-parametric part in the first-stage with a second stage estimation of the treatment effect based on Neyman-orthogonal moment equations for robust estimation of treatment effects. DML procedure can achieve a $\sqrt{n}$-consistent estimator of the treatment effect under very weak conditions. \cite{semenova2017estimation} extend DML to a panel data setting with a high-dimensional treatment effect and unobserved unit heterogeneity, but limit the relationship between features and treatment/outcome to be linear. Other extensions and variations of the DML framework are also available in the literature \cite{mackey2018orthogonal,oprescu2019orthogonal}. Nevertheless, DML methods and its variants have practically been mostly applicable to partially linear regression models or setups with binary treatment. \cite{nekipelov2022regularised} is a recent work that proposes an Orthogonal ML approach for nonlinear semi-parametric models and is applicable to our problem setup. 

Machine learning methods such as random forests and neural networks have recently been used to build methods for estimating treatment effects and counterfactual prediction. Some recent works have proposed methods for estimating non-parametric heterogeneous treatment effects based on random forests \cite{athey2019generalized,wager2018estimation}. Other works have employed neural networks for learning treatment invariant representation of confounders to adjust the observational distribution \cite{shalit2017estimating}, and variational autoencoders (VAE) \cite{kingma2013auto,rezende2014stochastic} for inferring unobserved confounders with proxies \cite{louizos2017causal} or inferring latent factors of treatment and decorrelating them with confounders \cite{zou2020counterfactual}. However, all of these works focus mainly on binary (or vector of binary) treatments. In contrast, our focus in this work is on a continuous treatment variable such as price with the response being modeled using discrete distributions such as Poisson with a non-linear demand-response function to model the price-demand relationship.

\subsection{Overview of the Paper and Main Contributions}

In this article, we consider a setting in which a firm can measure relevant confounders for estimating price-sensitivity. However, because the firm is dealing with a large number of features with complex interactions, fully parametric models for constructing demand response functions may not suffice. To address this situation, we formulate a Poisson semi-parametric demand response model with feature-dependent price-sensitivity. This model allows us to leverage the flexibility provided by modern ML techniques, while keeping the specification of the effect of price on demand interpretable. Pricing methodologies can have a large impact on the firm's revenue and for such critical tasks, interpretability vis-a-vis price sensitivity parameters is important as it allows a firm to qualitatively validate the inferred parameters easily by considering the individual effect of each feature. For example, parameters associated with day of week can be used to analyze relative impact of each day on price-sensitivity to see if weekdays have higher willingness-to-pay as compared to weekends on markets with predominant business traffic. Similarly, parameters associated with time of day can be checked to see if morning and evening flights have relatively higher willingness-to-pay as compared to afternoon flights on weekdays etc. We propose two approaches to construct estimators of price sensitivity parameters associated with this demand model.

The modeling and estimation aspects related to the Poisson semi-parametric model are not straightforward. To address that, we first develop a direct approach to implement this model using deep neural networks. The estimates for price-sensitivities can be directly obtained from the weights of one of the layers of the trained neural network. While this methodology is relatively easy to implement, it may be challenging to validate, thereby making it less feasible from a practical application perspective. Moreover, the estimates of price-sensitivities may be susceptible to bias since we simultaneously estimate the price-sensitivity parameters with other unknown parameters of the model and use regularization techniques to prevent overfitting.

To address some of the issues of the direct estimation approach described above, our main contribution involves development of a robust two-stage approach for estimating price-sensitivities by extending the ideas related to Double or Orthogonal machine learning \cite{chernozhukov2018double} to a Poisson semi-parametric setting. In the first stage of the method, we leverage the predictive power of modern machine learning methodologies to build predictive models for price and demand conditional on other features. In the second stage, we utilize interpretable parametric models and employ dynamic Bayesian models with an efficient sequential learning scheme based on variational Bayes ideas and Laplace approximation. The Bayesian model is naturally equipped with uncertainty estimation, allows for encoding prior information, and exhibits robustness in our experiments. In addition, it is suitable for streaming data and works well in the presence of time-variation of parameters that may exist in practice. 

We also discuss the related price optimization problem and how the uncertainty estimates related to price-sensitivities can be used for evaluating policies for price experimentation  based on Reinforcement Learning (RL) techniques such as Upper Confidence Bound (UCB) etc. \cite{Sutton2018}. 

To compare the performance of the proposed methodologies, we conduct a simulation study and show that the two-stage construction leads to robust and more accurate estimators of price-sensitivity parameters when compared to the direct one-stage approach. We also conduct a numerical study using real data to qualitatively compare the performance of the two estimation approaches.

\section{Modeling Perspective and Framework}\label{sec:framework}

\subsection{Poisson Semi-parametric Model for Demand Response}\label{sec:sp_framework}

In this section, we formally define the problem set-up in the dynamic pricing setting under consideration.  Consider a selling horizon of $T$ time periods, where for each $t = 1, 2, \ldots, T$, the seller offers an itinerary $i$ at price $P_{it}$. Assuming that customers arrive according to a Poisson process and considering an exponential form for the demand response model, the demand and observed prices are modeled as
\begin{eqnarray}
	Y_{it}|\bs{X_{it}},P_{it} & \sim &  \mathtt{Pois}(\lambda(P_{it}; \bs{X}_{it}, \bs{\theta}, \phi)), \quad \mathtt{log}(\lambda(P_{it}; \bs{X}_{it}, \bs{\theta}, \phi))= P_{it} \bs{\theta}^T\bs{W_{it}} + \phi(\bs{X_{it}}), \label{eq:main_sp}\\
	P_{it}|\bs{X_{it}} & = & g(\bs{X_{it}}) + \epsilon_{it}, \quad \mathbb{E}[\epsilon_{it}|\bs{X_{it}}]=0. \label{eq:aux_sp}
\end{eqnarray}

Here we denote the number of observed purchases for this itinerary in time period $t$ by $Y_{it}$. The vector of controls, denoted by $\bs{X_{it}}\in \mathbb{R}^m$, includes exogenous features like data related to the specific itinerary derived features or pre-specified transformations of this data (e.g., time seasonality basis functions, constant term etc.), and may also contain lags of $Y_{it}$ and $P_{it}$. $\bs{W_{it}} \in \mathbb{R}^d$ is the feature set related to the heterogeneity associated with the price-elasticity. $\bs{W_{it}}$ is based on a sub-vector of the observed data vector $\bs{X_{it}}$; it includes an intercept term and any pre-specified transformation of raw features. $\bs{\theta} \in \mathbb{R}^d$ is a vector of parameters associated with price-sensitivity. $\phi(\cdot)$ is a function associated with the total demand volume. To illustrate its role, note that if we set $P_{it} = 0$ in \eqref{eq:main_sp}, the purchases will occur according to a Poisson process with mean $\mathtt{exp}(\phi(\bs{X_{it}})))$. As discussed earlier, the overall demand for a product could be influenced by a variety of factors, therefore $\phi(\cdot)$ could be non-linear, complex and high-dimensional. Given this model, our main objective is to estimate the price-sensitivity parameters $\bs{\theta}$. 

In \eqref{eq:aux_sp}, we have specified the dependence of price on the set of features $\bs{X_{it}}$. Typically, with the RM and Pricing systems employed by the firm, the prices are based on marginal opportunity costs (bid prices) computed by solving a finite horizon optimization problem considering demand forecasts, capacities etc., and further refinement based on pricing rules employed by the firm. Therefore, the function $g(\cdot)$ in most practical scenarios will be complex and high-dimensional. Although \eqref{eq:aux_sp} is not the main equation of interest, it is useful in elucidating the causal dependence between variables in the model and will also help us later in developing a robust two-stage estimation approach. While our main concern is to estimate the price sensitivity parameters $\bs{\theta}$, we still need to estimate the additional unknown parameters of the model, $((\phi(\cdot), g(\cdot))$, known in the econometrics parlance as the nuisance parameters. As compared to a fully parametric model such as the Poisson generalized linear model, which itself may lead to biased estimates of $\bs{\theta}$ if the nuisance (volume) parameter is mis-specified due to linearity restriction, the semi-parametric model in \eqref{eq:main_sp} with a more flexible structure allows one to mitigate this problem while still being interpretable and amenable to efficient estimation approaches for the parameters of interest \cite{Robinson1988}.  

\section{Estimating Price-Sensitivity Parameters}\label{sec:estimation}
Estimating the parameters of the Poisson semi-parametric model described in \eqref{eq:main_sp} can in practice be challenging. Ideally, one wants $\bs{X_{it}}$ to incorporate all the possible features and confounders (to control for them), including possible interactions and transformations. To alleviate the challenges related to complex feature engineering, it is beneficial to utilize sophisticated non-parametric ML models in constructing the estimators for the nuisance parameters, $((\phi(\cdot), g(\cdot))$. We next propose two approaches based on these ideas for estimating the parameters of interest.

\subsection{Direct Estimation using Wide and Deep Neural Networks}\label{sec:wide_deep}
\begin{figure}[h]
	\centering
	\includegraphics[width=0.5\linewidth]{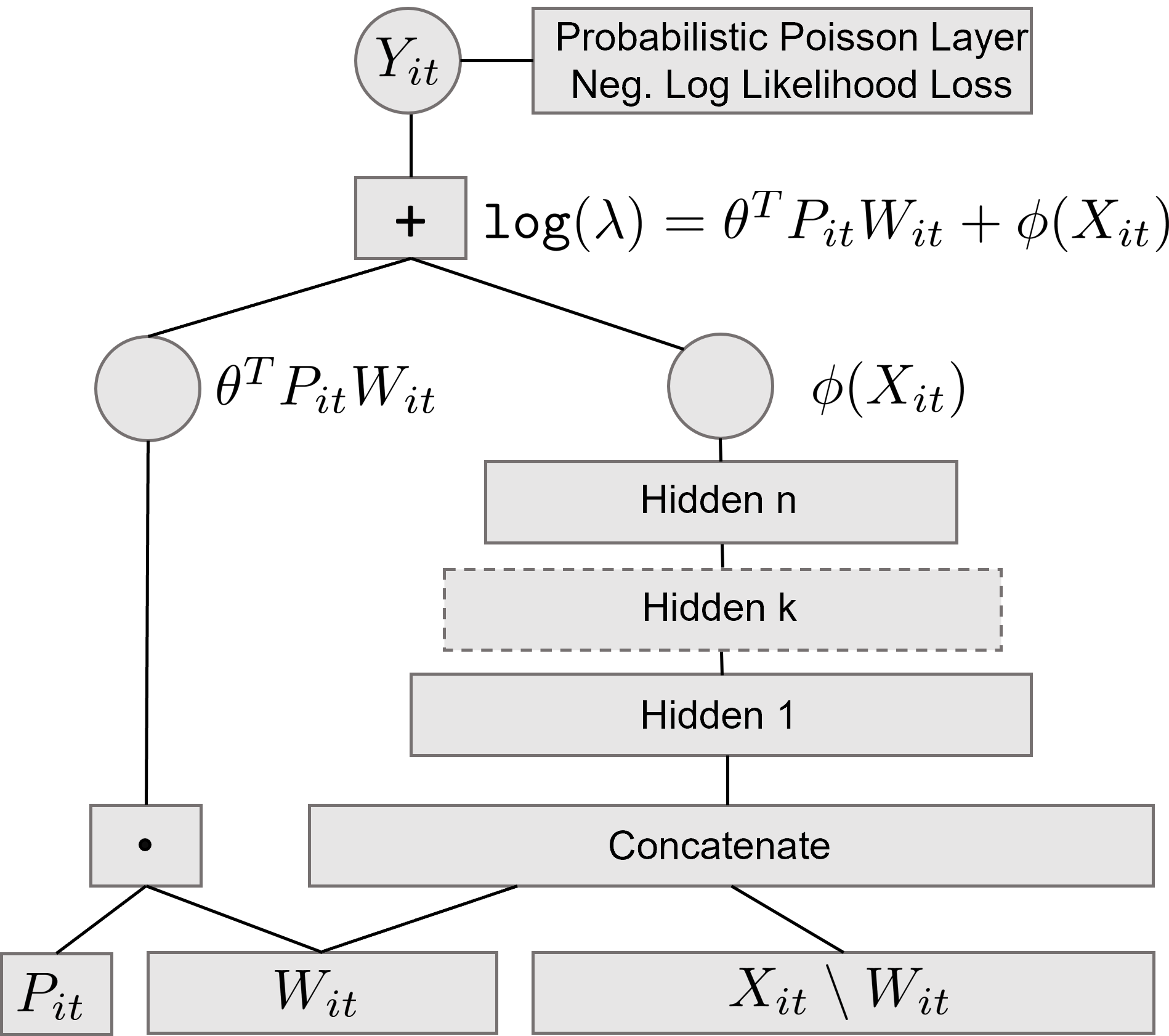}
	\caption{Wide \& Deep Neural Network Architecture for the Poisson Semi-parametric Model}
	\label{fig:wide_deep_nn}
\end{figure}

In the first approach, we model \eqref{eq:main_sp} using a deep neural network and directly estimate the parameters, $\bs{\theta}$ and $\phi(\cdot)$, by training it on the historical sales transaction data. We would like to leverage the advantages of complex feature representations learned via the deep neural network for estimating the nuisance parameter $\phi(\cdot)$. However, since the dependency of price sensitivity on the features is log-linear, we need to consider a specific neural network architecture for this purpose. The Wide \& Deep neural network architecture proposed by \cite{Cheng2016} provides a way to accomplish this by jointly training a wide linear part and a deep neural network. Although this architecture was originally proposed for recommender systems, we adapt it for the price-sensitivity estimation problem. Figure \ref{fig:wide_deep_nn} shows this architecture for the Poisson semi-parametric model in \eqref{eq:main_sp}. 

In this architecture, the wide component is used to model the linear part of the form $P_{it} \bs{\theta}^T\bs{W_{it}}$ and the deep part is used to model the nuisance parameter $\phi(\bs{X_{it}})$. These parts are combined to construct the log rate parameter and fed to a Probabilistic Poisson layer with Poisson negative log likelihood loss for joint training. The estimated parameters of interest, $\hat{\bs{\theta}}$, can be obtained from the weights of the layer constructing the log-rate parameter.   

This method provides a direct way to construct an estimator for the Poisson semi-parametric model and can be implemented relatively easily using packages such as {\tt tensorflow} or {\tt PyTorch}. However, in a practical implementation, this approach can be challenging to work with since the output of the deep part is the nuisance parameter, $\phi(\cdot)$, which is not directly observable. Therefore, we cannot validate how well the deep part is learning this parameter. Moreover, to control for overfitting, a regularization approach (e.g., L1/L2 norm penalization of model parameters, early stopping, drop-out or batch normalization) is often necessary while training the neural network. But regularization may also introduce bias in the estimate of the nuisance parameter. In the absence of properties that make the estimation of price sensitivity parameters robust to errors in the nuisance parameter, any bias in learning the nuisance parameter will lead to bias in the estimate of price-sensitivity parameters $\bs{\theta}$. 

An alternative approach to the Wide \& Deep architecture proposed above is to directly use a sophisticated ML estimator (e.g., deep neural networks or gradient boosted trees) for constructing the estimator for $\phi(\cdot)$ in \eqref{eq:main_sp} and estimate the price sensitivity parameters by an iterative scheme that alternates between estimating the parametric and non-parametric parts. We can start by initializing the estimator of $\phi(\cdot)$ to $\hat{\phi}_0(\cdot)$ e.g., by randomly initializing the weights of a neural network. Given the initial estimate of $\hat{\phi}_0(\cdot)$, we can estimate the price-sensitivity parameters, $\hat{\bs{\theta}}_0$,  by maximum likelihood estimation. Given the estimated  $\hat{\bs{\theta}}_0$ we can then create an updated estimate of $\phi(\cdot)$, $\hat{\phi}_1(\cdot)$. This process continues until some convergence criteria is met for the price-sensitivity parameter. However, this approach not only suffers from the regularization related issue described above but due to its iterative nature can also become prohibitively computationally expensive for practical implementation.

\subsection{Two-stage Estimation Approach}\label{sec:prop_estimation}

To address deficiencies of the estimation approaches discussed so far, we present a two-stage approach in this section. Given the data $\mathcal{D} = (Y, \bs{X}, P)$, the log-likelihood function for our original model in \eqref{eq:main_sp} can be written as
\begin{equation}
	\ell(\mathcal{D}; \bs{\theta}, \phi) = Y(P\bs{\theta}^T \bs{W}+\phi(\bs{X})) - \mathtt{exp}(P\bs{\theta}^T \bs{W} + \phi(\bs{X})) - \mathtt{log}(Y !).
\end{equation}

Let $\bs{\Theta} \subset \mathbb{R}^d$ be the convex feasible space for price-sensitivity parameters and $\Phi$ be the set of functions of $\bs{X}$ with finite mean squared error. The true values of the parameters of \eqref{eq:main_sp} are,
\begin{equation*}
	(\bs{\theta}_0, \phi_0) = \arg\max_{\bs{\theta\in\bs{\Theta}}, \phi \in \Phi} \mathbb{E}[\ell(\mathcal{D}; \bs{\theta}, \phi)],
\end{equation*} 
where the expectation is taken w.r.t. the probability law governing the data $\mathcal{D} = (Y, \bs{X}, P)$. Since the log-likelihood function is concave, it satisfies the first order optimality condition given the true parameters $(\bs{\theta}_0, \phi_0)$

\begin{equation*}
	{\frac{\partial}{\partial {\bs{\theta}}}\mathbb{E}[\ell(\mathcal{D}; \bs{\theta}, \phi)]}\Big|_{\bs{\theta}_0,\phi_0} = 0.
\end{equation*}

The score function, $\psi(\mathcal{D}; \bs{\theta}, \phi) = \frac{\partial}{\partial {\bs{\theta}}}\mathbb{E}[\ell(\mathcal{D}; \bs{\theta}, \phi)]$, for estimating $\bs{\theta}_0$ satisfies the Neyman orthogonality condition if the Gateaux derivative of the score function with respect to the the nuisance parameter $\phi$ at $\phi_0$ is 0 \cite{chernozhukov2018double, neyman1979c} i.e.,

\begin{equation}\label{eq:NeymanOrtho}
	\frac{\partial}{\partial r}\psi(\bs{\theta}_0, \phi_0 + r(\phi - \phi_0))\Big|_{r=0} = 0.
\end{equation}

The condition in \ref{eq:NeymanOrtho} intuitively means that the score function for estimating the parameter of interest, $\bs{\theta}$, is locally insensitive to small perturbations in the nuisance parameter around $\phi_0$. However, the score function, $\psi(\mathcal{D}; \bs{\theta}, \phi)$, corresponding to our original Poisson semi-parametric model in \eqref{eq:main_sp} doesn't satisfy this orthogonality condition. Thus, we next describe a two-stage estimation method which leads to a \textit{reduced form} model with score function that is indeed Neyman orthogonal.

\subsubsection{Constructing Neyman Orthogonal Score Functions.}\label{sec:neyman_derivation}
In \cite{chernozhukov2018double} the authors proposed a two-stage estimation approach where in the first stage the nuisance parameters are estimated using sophisticated ML estimators and in the second stage the parameters of interest e.g., the price-sensitivity parameters in our case, are estimated. The second stage score function is constructed so that it satisfies the Neyman orthogonality condition \eqref{eq:NeymanOrtho}. Most of the works on Neyman orthogonal scores e.g., \cite{chernozhukov2018double, semenova2017estimation},  consider linear or log-linear models with additive noise. For the Poisson model in our case with non-linear demand response, we construct the Neyman orthogonal scores approximately via the Concentrating-out approach proposed in \cite{chernozhukov2018double}. This approach works on the principle that maximizing the expected maximum likelihood function with respect to the nuisance parameter, $\phi$, plugging that maximizer back in, and differentiating it with respect to the parameter of interest $\mathbf{\theta}$ produces an orthogonal moment condition.

For this let 
\begin{equation*}
	\phi_{\bs{\theta}} = \arg\max_{\phi \in \Phi} \mathbb{E}[\ell(\mathcal{D}; \bs{\theta}, \phi)], 
\end{equation*} 
be the ``concentrated-out'' non-parametric part of the model. Note that there is a $\phi_{\bs{\theta}}$ for any $\bs{\theta} \in \bs{\Theta}$. 

For our model, this can be derived based on first order conditions as  

\begin{equation}
	\phi_{\bs{\theta}}(\bs{X}) = \mathtt{log}(\mathbb{E}[Y|\bs{X}]) - \mathtt{log}(\mathbb{E}[ \mathtt{exp}(P\bs{\theta}^T \bs{W})|\bs{X}]) \approx \mathtt{log}(\mathbb{E}[Y|\bs{X}]) - \bs{\theta}^T W\mathbb{E}[P|\bs{X}], \bs{\theta} \in \bs{\Theta},
	\label{eq:conc_nuisance}
\end{equation} 
where the approximation is based on a first order Taylor series expansion of the exponential function around $\mathbb{E}[P|\bs{X}]$ (details in Section \ref{sec:Appendix1} of the Online Appendix). For the price-sensitivity estimation task we are interested in, this approximation works reasonably well in practice where the term $(\bs{\theta}^T W\sigma_{\epsilon_{it}})^2$ is small. 

Based on the approximate ``concentrated-out'' non-parametric part, we can construct an alternate score function for estimating $\bs{\theta}$ which is Neyman orthogonal \cite{chernozhukov2018double} as 
\begin{equation}
	\bar{\psi}(\mathcal{D}; \bs{\theta}, \phi) = \frac{d\ell(\mathcal{D}; \bs{\theta}, \phi_{\bs{\theta}})}{d\bs{\theta}}.
	\label{eq:conc_score}
\end{equation}
For the Poisson semi-parameteric model, based on \eqref{eq:conc_nuisance} and \eqref{eq:conc_score}, the Neyman orthogonal score is given by
\begin{eqnarray} \label{eq:neyman_ortho_score}
	\bar{\psi}(\mathcal{D}; \bs{\theta}, \phi_{\bs{\theta}}) = Y\bs{W}(P - \mathbb{E}[P|\bs{X}]) - \mathtt{exp}\Big(\bs{\theta}^T\bs{W}(P - \mathbb{E}[P|\bs{X}])+\mathtt{log}(\mathbb{E}[Y|\bs{X}])\Big)\bs{W}(P - \mathbb{E}[P|\bs{X}]).
\end{eqnarray}
Note that the Neyman orthogonal score function for estimating $\bs{\theta}$ in \eqref{eq:neyman_ortho_score} corresponds to the \textit{reduced form} model  
\begin{eqnarray}
	Y_{it}|\bs{X_{it}},P_{it} & \sim &  \mathtt{Pois}\big(\mathtt{exp}((P_{it} - \mathbb{E}[P_{it}|\bs{X_{it}}])  \bs{\theta}^T\bs{W_{it}} + \mathtt{log}(\mathbb{E}[Y_{it}|\bs{X_{it}}]))\big). \label{eq:main_sp_local}
\end{eqnarray}
Therefore, instead of the original model, we can estimate the price-sensitivity parameters based on the \textit{reduced form} equation \eqref{eq:main_sp_local}. The subtraction of $\mathbb{E}[P_{it}|\bs{X_{it}}]$ from $P_{it}$ in \eqref{eq:main_sp_local} changes the price-related variable
to have a ``local" interpretation, in which the parametric part locally governs changes in demand w.r.t. changes in price around the expected value for price given $\bs{X_{it}}$.

\subsubsection{Advantages of the Reduced Form Model.}
Working with the \textit{reduced form} version of the demand response model is advantageous for several reasons. Firstly, instead of the nuisance parameters $\phi(\cdot)$ and $g(\cdot)$ in \eqref{eq:main_sp}, the nuisance parameters in \eqref{eq:main_sp_local} are expected purchases $\mathbb{E}[{Y}_{it}|\bs{X}_{it}]$ and expected price $\mathbb{E}[{P}_{it}|\bs{X}_{it}]$. These are directly observable quantities so we can deploy sophisticated ML techniques for constructing estimators $\hat{Y}_{it}$ and $\hat{P}_{it}$ for $\mathbb{E}[{Y}_{it}|\bs{X}_{it}]$ and $\mathbb{E}[{P}_{it}|\bs{X}_{it}]$; validation of their goodness of fit becomes much easier, enabling us to use standard techniques for model selection and hyperparameter tuning. Furthermore, learning $\hat{P}_{it}$ is akin to learning the firm's historical pricing policy which can be useful in constructing reasonable price bounds for an automated dynamic pricing system in a practical deployment. Secondly, given estimates $\hat{P}_{it}$ and $\hat{Y}_{it}$, the model in \eqref{eq:main_sp_local} is now a fully parametric Poisson generalized linear model, therefore it is easy to use batch or online estimators for estimating the price sensitivity parameters $\bs{\theta}$. Finally, and most importantly, by approximately orthogonalizing the price with respect to $\bs{X}$ and removing the effect of confounding by subtracting $\hat{P}_{it}$ from $P_{it}$ in \eqref{eq:main_sp_local}, the \textit{reduced form} model provides robustness in the subsequent estimation of parameters $\bs{\theta}$ against biases in the estimators for nuisance parameters \cite{chernozhukov2018double}.

\subsubsection{Two-stage Approach based on the Reduced Form.} \label{sec:two-stage-outline}
Our overall scheme for constructing the estimator of price-sensitivity parameters $\bs{\theta}$  based on the \textit{reduced form} \eqref{eq:main_sp_local} comprises of two stages. In the first-stage, we create two estimators, $\hat{P}_{it}$ and $\hat{Y}_{it}$, for $g(\bs{X_{it}}) = \mathbb{E}[P_{it}|\bs{X_{it}}]$ and $\mathbb{E}[Y_{it}|\bs{X_{it}}]$, respectively, using machine learning methods. For $\hat{P}_{it}$ the response variable is the observed price and for $\hat{Y}_{it}$ the response variable is the observed bookings. Given the estimates from the first stage, in the second stage we estimate parameters of a Poisson generalized linear model. The use of non-parametric ML-based methods for the estimators of $\hat{P}_{it}$ and $\hat{Y}_{it}$ allows us to bring more complex representation of the feature space $\bs{X_{it}}$ into the model while still being able to perform the second-stage estimation via a tractable and interpretable Poisson generalized linear model form.


In this work, we use neural networks for constructing the first-stage estimators and Bayesian dynamic generalized linear models (DGLM) for the second stage estimation as will be discussed in the following sections. The use of Bayesian methods in the second stage is mainly done to get online estimators of $\bs{\theta}$ which are convenient for practical applications. Any concern regarding the regularization bias due to implicit regularization in Bayesian methods can be mitigated by considering prior distributions with high variance. Furthermore, the Bayesian approach also provides a principled way to quantify uncertainty on the estimate of price-sensitivity parameters which is not only useful for validating the model but can also drive reinforcement learning-based approaches for price optimization. Similar to \cite{chernozhukov2018double, semenova2017estimation, athey2019generalized}, and \cite{mackey2018orthogonal}, we use cross-fitting approach in our two-stage scheme to prevent overfitting, where the data is split into $K$-folds and for each fold the first stage predictions used in the second stage estimation are from first-stage models trained on other folds. As shown in \cite{chernozhukov2018double} prevention of over-fitting in the construction of first stage estimators for $\mathbb{E}[Y_{it}|\bs{X_{it}}]$ and $\mathbb{E}[P_{it}|\bs{X_{it}}]$ plays a key role in getting $\sqrt{n}$-consistent two-stage estimators for $\bs{\theta}$.

In \eqref{eq:main_sp}, we assume a parametric form for treatment heterogeneity because of its interpretability. In practice, a domain expert may choose $\bs{W_{it}}$ considering the trade-off between flexibility and estimation precision. $\bs{W_{it}}$ can, for example, be a vector of binary variables encoding categorical features and other continuous features such as basis functions to capture seasonality etc. In our simulation experiments, we empirically see that our method provides elasticity estimates that are robust to minor misspecification in $\bs{W_{it}}$ when compared to simpler approaches. We will refer to this two-stage estimation approach as \texttt{Two-stage-CO} in the rest of this article.

\subsubsection{Neural-networks based First Stage Prediction of Price and Bookings.}

For real-world applications, we use neural networks to construct the estimator for $\mathbb{E}[P_{it}|\bs{X_{it}}]$ in the first stage with the mean-squared error loss. Although in principle we can use other function approximators such as gradient boosting, neural networks can provide advantages beyond just the prediction task. One such advantage is the ability to encode categorical variables in our feature set via learnable embedding vectors. We can cluster categorical variables based on the closeness (euclidean norm) of embedding vectors e.g., derive groups or segments related to Origin-Destination pairs, Days of Weeks etc., with similar pricing behavior. Such inferred closeness of categories contains information about expert knowledge encoded in the pricing system, which then can be used to inform the choice of the structure of $\bs{W}$ and/or segmentation to help with data sparsity.

When learning the first stage model for price, we leverage all the available data from different markets (Origin-Destination pairs) and add categorical variables representing each market to the feature set. This also requires careful scaling of the prices in different markets to a similar range so that they have a balanced influence on the training. By pooling all the data, we increase the sample size and also effectively add an implicit regularization to learn a more accurate model compared with training on each market separately. In addition, we embed the categorical features in a low-dimensional continuous space, where the embedding is learned jointly with other parameters of the neural network. 

We also estimate $\mathbb{E}[Y_{it}|\bs{X_{it}}]$ using neural networks in the first stage. However, as the generating process for purchases is assumed be a count process, we train the model with a Poisson log-likelihood loss. Note that the choice of this loss function is still an approximation as although $Y_{it}|\bs{X_{it}}, P_{it}$ has a Poisson distribution, the distribution of $Y_{it}|\bs{X_{it}}$ is not Poisson. This approximation is mainly driven by the ease of implementing neural networks using standard packages such as \texttt{Pytorch} and \texttt{tensorflow}. Our numerical results in Sections \ref{sec:simulation} and \ref{sec:case_study} show that despite this approximation, the two-stage estimation approach performs well.

\subsubsection{Bayesian Dynamic Generalized Linear Model for Second Stage Estimation of Price Sensitivity Parameters.} \label{sec:inference}

Given the predictions of the first-stage models, in the second stage, we adopt a Bayesian dynamic generalized linear model (DGLM) \cite{west2006bayesian,berry2020bayesian} for estimating the price-sensitivity parameter $\mathbf{\theta}$ in \eqref{eq:conc_nuisance}. DGLMs are based on the Bayesian state-space models with a sequential and efficient update procedure, which makes them suitable for applications with streaming data. Additionally, the Bayesian framework provides a principled way for quantifying uncertainty of parameter estimates, which is helpful for making more informed pricing decisions and utilizing RL techniques. Based on the estimates from the first stage, the Bayesian model for the second stage for \eqref{eq:conc_nuisance} is:
\begin{eqnarray}
	Y_{it} &\sim&  \mathtt{Pois}(\lambda_{it}),\nonumber\\ \mathtt{log}(\lambda_{it})&=&(P_{it}-\hat{P}_{it}) \bs{\theta}^{T}\bs{W_{it}} + \mathtt{log}(\hat{Y}_{it})\quad \bs{\theta} \sim \pi(\bs{\theta}). \label{eq:main-latent}
\end{eqnarray}
Here, $\hat{Y}_{it}$ is the cross-fitted prediction of purchases for the specific itinerary $i$ from the first stage model and $\pi(\bs{\theta})$ is the prior distribution over the parameters. The inference in the second stage is done independently for each market on all itineraries belonging to that market. The inference procedure of the second stage can be completely parallelized for efficiency, which is extremely relevant for an   implementation in a real-life system. For each market, $\bs{W}_{it}$ is concatenation of an intercept, continuous features (e.g., Fourier basis for departure date seasonality, Splines related to days before departure, departure time etc) and one-hot encoded categorical variables (e.g., Day of Week, Point of Sale, Number of Connections etc.) as shown in \eqref{eq:W}.
\begin{equation} \label{eq:W}
	\bs{W}_{it}=[1,\underbrace{0.5,0.1,\cdots,0.7}_\text{Cont. Vars.},\underbrace{0,0,\cdots,1}_\text{Cat. Var. 1},\underbrace{0,\cdots,1,0}_\text{Cat. Var. 2},\cdots,\underbrace{0,1,0,\cdots}_\text{Cat. Var. v}]^T.
\end{equation}
Thus, by modeling heterogeneous price-sensitivity through an additive structure, the parameter $\bs{\theta}$ in \eqref{eq:main-latent} is estimated for each market (origin-destination pair) using data for all its departure dates present in the historical transactions.  


The update procedure in DGLMs is sequential and can be performed immediately after each new observation 
(daily bookings and price recorded for a market). The overall procedure is based on the combination of linear Bayes' theory \cite{goldstein2007bayes} and a variational Bayes approach. We describe the inference procedure in detail in Section \ref{sec:Appendix2} of the Online Appendix, where we also introduce a Laplace approximation based approach to further speed up the procedures of \cite{berry2019bayesian} and \cite{west2006bayesian}. Through this inference procedure, we are able to update the posterior means and covariances of the parameters without requiring any assumptions on their exact distribution. 

A desirable property of the dynamic regression model is that it can be incrementally updated in an efficient manner in between the complete retraining of the first and second stage models. In our application, the updating schedule of the first stage and second stage models are different. Since the airlines pricing policies and the underlying model of latent variables are expected to be more stable, we update the first stage models, which have higher computational cost, with lower frequency compared with the second stage model. The second stage model is incrementally updated after each observation to have the most updated belief on $\bs{\theta}$. 

\subsubsection{Alternative Approach to Neyman Orthogonal Scores via Partialling-out.}\label{sec:semenova}
An alternative approach to construct Neyman orthogonal scores for semi-parametric generalized linear models has recently been proposed in \cite{nekipelov2022regularised}. In this approach, the authors first partial out \cite{Robinson1988} the effect of controls ($\bs{X}$)  from treatment ($P$) inside the link function's argument of the GLM. 
Specifically, for the case of Poisson semi-parametric model \eqref{eq:main_sp}, in this approach one starts with the following partialled-out form for the original model \eqref{eq:main_sp}: 
\begin{eqnarray}
	Y_{it}|\bs{X_{it}},P_{it} & \sim &  \mathtt{Pois}\big(\mathtt{exp}((P_{it} - \mathbb{E}[P_{it}|\bs{X}_{it}])  \bs{\theta}^T\bs{W_{it}} + q(\bs{X}_{it}))\big). \label{eq:semenova_model}
\end{eqnarray}
where $q(\bs{X}_{it}) := \mathbb{E}[P_{it}|\bs{X}_{it}]\bs{\theta}_0^T\bs{W_{it}} + \phi_0(\bs{X}_{it}) = \mathbb{E}[\mathtt{log}(\mathbb{E}[Y_{it}|\bs{X_{it}},P_{it}])|\bs{X}_{it}]$. The Neyman orthogonal score for the Poisson semi-parametric model based on \cite{nekipelov2022regularised} is:
\begin{eqnarray}
	\tilde{\psi}(\mathcal{D}; \bs{\theta}, q) = \frac{Y\bs{W}(P - \mathbb{E}[P|\bs{X}]) - \mathtt{exp}\Big(\bs{\theta}^T\bs{W}(P - \mathbb{E}[P|\bs{X}])+q(\bs{X})\Big)\bs{W}(P - \mathbb{E}[P|\bs{X}])}{\mathbb{E}[Y|\bs{X},P]}, \label{eq:semenova_score}
\end{eqnarray}
which corresponds to a weighted maximum likelihood based score function for \eqref{eq:semenova_model} with the weights being $\frac{1}{\mathbb{E}[Y|\bs{X},P]}$. 

To utilize this score function, we first need to construct (ML) estimators for $\mathbb{E}[P_{it}|\bs{X}_{it}]$ and $\mathbb{E}[Y_{it}|\bs{X}_{it},P_{it}]$. The predictions from the initial estimator for $\mathbb{E}[Y_{it}|\bs{X}_{it},P_{it}]$, are used as weights and also as response variables to form an estimator for $q(\bs{X}_{it})$, i.e. an estimator for $\mathbb{E}[\mathtt{log}(\mathbb{E}[Y_{it}|\bs{X_{it}},P_{it}])|\bs{X}_{it}]$. After constructing these estimators, the parameter of interest, $\bs{\theta}$, is found based on the score function in \eqref{eq:semenova_score}. Given the first-stage estimators, the solution to \eqref{eq:semenova_score} can be easily found by fitting a regular weighted Poisson generalized linear model. We will refer to this partialled-out approach as \texttt{Two-stage-PO} and the concentrated-out approach in Section \ref{sec:neyman_derivation} as \texttt{Two-stage-CO}. 
It is worth noting that the partialling-out approach involves building (ML) estimators for $\mathbb{E}[P_{it}|\bs{X}_{it}]$, $\mathbb{E}[Y_{it}|\bs{X}_{it},P_{it}]$, and $q(\bs{X}_{it})$, while the \texttt{Two-stage-CO} approach only needs (ML) estimators for $\mathbb{E}[P_{it}|\bs{X}_{it}]$ and $\mathbb{E}[Y_{it}|\bs{X}_{it}]$. Thus, the  \texttt{Two-stage-CO} method can be significantly better in terms of computation time. Moreover, the inclusion of an additional estimator adds to the complexity of the process and requires increased effort in validating the goodness of fit of these estimators. We will compare the estimation performance of \texttt{Two-stage-CO} and \texttt{Two-stage-PO} via Simulation studies in Section \ref{sec:simulation}. 
\section{Price Optimization}






In this section we describe how to compute dynamic pricing policy given the estimates of the price-sensitivity parameters. The objective of the firm is to maximize the expected margin contribution given the associated per unit cost $c$ from selling the product to a request during a shopping session. As noted earlier, for the airline pricing problem, the cost is the marginal opportunity cost or bid price associated with the seat inventory requested in the itinerary. Here we assume that the applicable marginal opportunity cost has already been computed by the airline's RM system by solving a separate finite horizon optimization problem and is available at the time of the request. 

Let $\bs{\mu}_{\theta}$ and $\Sigma_{\theta}$ be the current estimate of mean and covariance associated with the price-sensitivity parameters $\bs{\theta}$ at the time of request, and $\mathcal{P}_{it} := p^{lb}_{it}\leq p \leq p^{ub}_{it}$ be the feasible space of prices for the request. Let $R_{it}(p; c, \bs{X}_{it}, \bs{\mu}_{\theta},\Sigma_{\theta})$ denote the expected margin contribution from an itinerary/product $i$ requested at time $t$ with associated feature vector $\bs{X}_{it}$ and cost $c$. The optimal price for the request is given by 
\begin{eqnarray} 
	P_{it}^{\ast} &=& \arg\max_{p \in \mathcal{P}_{it}} R_{it}(p; c, \bs{X}_{it}, \bs{\mu}_{\theta},\Sigma_{\theta}).\label{eq:price_opt_}
\end{eqnarray}
In Section \ref{sec:greedy} we discuss how we compute the dynamic prices in our experiments. 

\subsection{Computing Dynamic Prices} \label{sec:greedy}
Given the posterior distribution of price sensitivity parameter, $\pi^{*}(\bs{\theta})$, the expected margin contribution for the Poisson demand response model in \eqref{eq:main_sp} is:
\begin{equation}
	R_{it}(p; c, \bs{X}_{it}, \bs{\mu}_{\theta}, \Sigma_{\theta}) = \mathbb{E}_{\pi^{*}(\bs{\theta})}[ \mathtt{exp}(p\bs{\theta}^T\bs{W}_{it})(p-c)].\label{eq:bay-rew_}
\end{equation}
Finding the price based on maximizing the above expected margin contribution is akin to using a greedy policy for pricing. Since the Bayesian DGLM-based update procedure in \ref{sec:inference} does not make distributional assumptions on $\pi^{*}(\bs{\theta})$ and only yields the posterior mean and covariance of $\bs{\theta}$, a closed-form expression of \eqref{eq:bay-rew_} is not available, and one can either resort to a plug-in approach using point estimates of the parameters or an approximation of the expectation. We discuss the plug-in approach that we use in our experiments in the following and two more approaches to approximate \eqref{eq:bay-rew_} in Section \ref{sec: bayes_greedy} of the Online Appendix.

Using the posterior mean of the parameters, which is the optimal mean-squared error estimator, we can form a plug-in estimator of \eqref{eq:bay-rew_}, where the corresponding margin contribution function is
\begin{equation}
	R_{it}(p; c, \bs{X}_{it}, \bs{\mu}_{\theta}) = \mathtt{exp}(p\bs{\mu}_{\theta}^T\bs{W}_{it})(p-c).\label{eq:plugin_}
\end{equation} 
The utility function in \eqref{eq:plugin_} results in a convex optimization problem that in fact has a closed-form solution
\begin{equation}
	P_{it}^{\ast} = \min\{\max\{p^{lb}_{it}, c - \frac{1}{\bs{\mu}_{\theta}^T\bs{W}_{it}}\}, p^{ub}_{it}\}.\label{eq:plug-sol_}
\end{equation} 
Note that if we employ frequentist methods for estimating $\bs{\theta}$ e.g., the direct estimation approach in Section \ref{sec:wide_deep}, we can use the same formula in \eqref{eq:plug-sol_} to compute the optimal price which maximizes the margin contribution by replacing $\bs{\mu}_{\theta}$ with the frequentist estimate $\hat{\bs{\theta}}$. 

Some other methods for constructing pricing policies such as Bayes greedy and Reinforcement Learning are discussed further in Section \ref{sec:price-opt} of the Online Appendix.

\section{Simulation Experiments}\label{sec:simulation}

In this section we present results from computational experiments on randomly generated data where we know the true ground demand model, i.e., we know the underlying price-demand relationship. The goal of this study is to analyze and compare the accuracy of the proposed estimators of price-sensitivity parameters based on simulated data.  

\subsection{Simple Example}
We first perform a comparison on a simple setup with randomly generated data where we set the ground truth of the parameters. By comparing the estimates with the true parameters, we can evaluate the performance of our proposed estimation methods. For this simulation, the exogenous features, $\bs{X} \in  \mathbb{R}^{10}$, are normally distributed, $\bs{X}_i \sim \mathcal{N}(0,\Sigma(\rho))$, where $\Sigma_{jk}(\rho)=\rho^{|j-k|}$ and $\rho$ is set to 0.5. The dependence of price on the features is assumed to be linear and the demand data are generated according to a Poisson model:
\begin{equation*}
	\begin{split}
		Y_i|\bs{X}_i,P_i & \sim \mathtt{Pois}\big(\mathtt{exp}(P_{i}(\theta_0 + \sum_{j=1}^{4}\theta_j X_{ij}) + \tau + \bs{\eta}^T[\bs{X}_i^T,X^2_{i1},X_{i2}X_{i3},X_{i3}X_{i4},X_{i4}X_{i5}]^T )\big)\\
		P_i & = 50 + \bs{\gamma}^T\bs{X}_i + \epsilon_i, \quad \epsilon_i \sim \mathcal{N}(0,\sigma=9).
	\end{split}
\end{equation*}
Here, $\tau=1.2$, $\bs{\eta} \in \mathbb{R}^{14}$ with all elements equal to 0.1, and $\bs{\gamma} \in \mathbb{R}^{10}$ with all elements equal to 3. As can be seen, the nuisance $\phi(\bs{X_i})$ is a simple polynomial of degree 2. The vector of parameters of interest, $\bs{\theta}^T=[\theta_0,\theta_1,\theta_2,\theta_3,\theta_4]$, is set as [-0.02,-0.005,-0.005,-0.005,-0.005]. 

We compare the performance of the direct estimation approach based on Wide and Deep architecture presented in Section \ref{sec:wide_deep}, and the two-stage estimation approaches discussed in Section \ref{sec:estimation}. For this example, the deep part of the Wide and Deep consists of one hidden layer with 50 units and ReLU activation followed by a layer with a single unit representing the nuisance parameter. Additionally, we use L2 regularization on hidden layer weights of the deep part, and in each run, we do early stopping and restore the best performing model state on a validation set, formed by randomly setting aside 20\% of training data. These are all techniques commonly used when training neural networks on real data which can introduce some level of bias in parameter estimates. The model is trained using the Adam optimizer \cite{kingma2015adam}. 

For the proposed two-stage methods, the \textit{reduced form} model corresponding to the \texttt{Two-stage-CO} approach in Section \ref{sec:neyman_derivation} is:
\begin{equation*}
	\mathtt{log}(\mathbb{E}[Y_{i}|\bs{X}_{i},P_{i}])  =  (P_{i} - \mathbb{E}[P_{i}|\bs{X_{i}}])(\theta_0 + \sum_{j=1}^{4}\theta_j X_{ij}) 
	+ \mathtt{log}(\mathbb{E}[Y_{i}|\bs{X_{i}}]). 
\end{equation*}
We use ridge regression for estimating $\mathbb{E}[P_{i}|\bs{X_{i}}]$ in the first stage. The first stage estimator for $\mathbb{E}[Y_{i}|\bs{X_{i}}]$ is constructed using a random forest regressor with mean-squared-error loss. All first stage model training and predictions are carried out by 5-fold cross-fitting. In this toy example, to isolate the effect of the reduced form model on estimation accuracy, we employ a frequentist Maximum Likelihood Estimation (MLE) approach for the Poisson GLM  in the second stage estimation of the parameter of interest, $\bs{\theta}$.

For the alternative \texttt{Two-stage-PO} estimation approach discussed in Section \ref{sec:semenova}, in the first stage, we first fit $\mathbb{E}[Y_{i}|\bs{X_{i}},P_i]$ with a random forest regressor with Poisson deviance loss, and then construct an estimator for $\mathbb{E}[\mathtt{log}(\mathbb{E}[Y_{i}|\bs{X_{i}},P_i])|\bs{X_{i}}]$ by training a random forest regressor with mean-squared-error loss on $\mathtt{log}$ of predictions of the previous random forest. Note that these steps are performed with 5-fold cross-fitting. To estimate the parameter of interest, in the second stage we use a frequentist weighted MLE approach for the Poisson GLM where samples are given weights inversely proportional to their first stage prediction of $\mathbb{E}[Y_{i}|\bs{X_{i}},P_i]$. 

We generate 10,000 samples from the assumed data generation model in each simulation run, and test the estimation accuracy of different methods on the generated data. The procedure of data generation and model testing is repeated ten times and the average results are shown in Table \ref{tab:toy}. 

\begin{table}[!h]
	\centering
	\begin{tabular}{|c|c|c|c|}
		\hline
		Method & Direct & Two-stage-CO & Two-stage-PO \\
		\hline 
		Estimation MAE & 0.00369 (0.00071) & 0.00115 (0.00042) & 0.00117 (0.00034)\\
		\hline
	\end{tabular}
	\caption{Average and standard deviation (in parenthesis) of mean absolute error (MAE) of estimates of $\bs{\theta}$ for direct and two-stage methods.}
	\label{tab:toy}
\end{table}

The two-stage methods result in significant improvement over the direct estimation procedure. Moreover, \texttt{Two-stage-CO} and \texttt{Two-stage-PO} show very close performance and the difference between them is not statistically significant.

\subsection{Simulation in Airline Setting}
To generate the sales transactions mimicking the complexities involved in real-life airline pricing, we model the interactions between RM and pricing with finite seat capacity. We set-up the simulation experiment considering a single-leg with a seat capacity of 100. The booking horizon for each departure date consists of 365 days grouped into 10 contiguous time periods called time frames (TF), such that the demand arrival rate within each TF is constant but varies across TFs. TF 0 is the furthest away from departure of a flight and TF 9 is closest to departure. The demand arrival is Poisson distributed with a rate that varies by week of year (WOY), booking period (TF), days of week (DOW) and point of sale (POS). Moreover, customer willingness-to-pay is exponentially distributed ($\mathtt{WTP}|\theta \sim \mathtt{exp(-\frac{1}{\theta})}$) and the price-sensitivity parameters for customers ($\theta$) vary by POS and TF. With this setup, the ground demand response model governing demand on day $t$ of the booking horizon for departure date $i$ is
\begin{equation*}
	Y_{it}|\bs{X}_{it}=(\mathtt{WOY, DOW, POS, TF}),P_{it}  \quad \sim   \mathtt{Pois}\big(\mathtt{exp}((P_{it}   \sum_{POS, TF}\bs{\theta}_{POS, TF}\mathbbm{1}_{POS, TF} + \phi(\bs{X}_{it})\big). \label{eq:dgp}
\end{equation*}

The true price-sensitivity parameters are shown in Table \ref{tab:alpha_estimates}. Note that as discussed previously, for the exponential demand response function, the optimal price given price sensitivity parameter $\theta$ and cost $c$ is given by 
\begin{equation*}
	p^{\ast}|\theta, c =\arg\max\{\mathtt{exp}(p\theta)(p-c)\} = c - \frac{1}{\theta}.
\end{equation*} 
For a given POS, TF combination, $\alpha = -\frac{1}{\theta}$ denotes the mean willingness-to-pay, which is also the optimal price when the cost is zero. These provide a more intuitive explanation of the price sensitivity parameters and therefore we have included them in the Table \ref{tab:alpha_estimates}. 

The RM system computes the marginal opportunity cost of seats (bid prices) considering the known demand arrival rates and price-sensitivity parameters using dynamic programming and the pricing system computes the final prices based on the relevant bid price and true price-sensitivity parameters i.e., the optimal pricing policy. To mimic idiosyncratic noise in prices observed in real life, we add normally distributed noise to final prices with mean zero and standard deviation of 20. These prices are then offered to a stream of arriving customers and the customer accepts or rejects the price based on their sampled willingness-to-pay ($\mathtt{WTP}|\theta \sim \mathtt{exp(-\frac{1}{\theta})}$). The number of customers arriving in any time unit and their WTP are sampled according to the encoded ground truth price-demand model. This process creates the historical sales transaction data which includes daily bookings, daily average price and additional features useful for estimation namely, $\mathtt{WOY, DOW, POS, TF}$. Given this scheme, the price generation process in the simulation is:
\begin{eqnarray*}
	P_{it}|\bs{X}_{it}=(\mathtt{WOY, DOW, POS, TF}) &=& v_{it}(s_{it}; \bs{X}_{it}) - \frac{1}{\theta_{POS, TF}} + \epsilon_{it}, \quad \epsilon_{it}|\bs{X}_{it}\sim \mathcal{N}(0,20),
\end{eqnarray*}
where $v_{it}(x)$ denotes the optimal opportunity cost (bid price) computed via dynamic programming based optimization for the departure date $i$ on day $t$ given that we have  $s_{it}$ seats remaining. Given the complex dependence of price on bid price, reservation process and price elasticity, the data generation process for bookings and prices resembles the model described in \eqref{eq:main_sp} and \eqref{eq:aux_sp}.

Using this simulation setup, we generate data corresponding to two years worth of daily flight departures (730 individual flights). Further analysis on the data generated under this simulation set-up can be found in the Online Appendix, Section \ref{sec:Appendix3}.

\subsubsection{Price Sensitivity Parameter Estimation.}
For the purpose of estimating the price-sensitivity parameters, we consider the following model for the log-rate of the Poisson demand response model
\begin{equation}
	\mathtt{log}(\mathbb{E}[Y_{it}|\bs{X}_{it},P_{it}])  =  P_{it}(\theta_0+\sum_{pos=1}^{\#POS-1}\theta_{pos}\mathbbm{1}_{\{POS=pos\}} + \sum_{tf=1}^{\#TF-1}\theta_{tf}\mathbbm{1}_{\{TF=tf\}}) + \phi(\bs{X}_{it}). 
	\label{eq:estimation_model}
\end{equation}

Note that the ground demand response model used for the data generating process has a weak interaction in the price sensitivity parameters between POS and TF. Given some of the sparser entities, estimation benefits from a simpler linearly additive specification of price effects across POS and TF as shown in \eqref{eq:estimation_model}. Moreover, minor misspecifications of this type are common in practice, so the model in \eqref{eq:estimation_model} will allow us to validate the robustness of the estimation method better. We use the direct neural network based technique and the two-stage approach to estimate the parameters of interest, $\bs{\theta}$, for the above model next.

\subsubsection{Direct Approach based on Wide and Deep Neural Network.}
We first perform the direct estimation method proposed in Section \ref{sec:wide_deep}. The features used in the model, $\bs{X}_{it}$, include one-hot encoded categorical variables for POS, TF, and DOW, and four continuous features representing the yearly seasonality, which are Fourier series based on WOY with the first two frequencies and 52 weeks period. The deep part consists of a hidden layer with 50 units and ReLU activation followed by a layer with a single unit. The layer constructing the Poisson log rate forms $P_{it}(\sum_{pos=1}^{\#POS-1}\bs{\theta}_{pos}\mathbbm{1}_{\{POS=pos\}}+ \sum_{tf=0}^{\#TF}\bs{\theta}_{tf}\mathbbm{1}_{\{TF=tf\}}) + \phi(\bs{X}_{it})$. In each run, we randomly set aside 15\% of flights as validation set and train the model using the Adam optimizer \cite{kingma2015adam} with early stopping. 
\subsubsection{Two-stage Approach.}

As detailed in Section \ref{sec:estimation}, we construct two-stage estimators for the price sensitivity parameters $\bs{\theta}$ using both \texttt{Two-stage-CO} and \texttt{Two-stage-PO} methods.

For the \texttt{Two-stage-CO} approach detailed in Section \ref{sec:neyman_derivation}, we use the following \textit{reduced form} model based on \eqref{eq:main_sp_local}
\begin{equation}
	\mathtt{log}(\mathbb{E}[Y_{it}|\bs{X}_{it},P_{it}])  =  (P_{it} - \mathbb{E}[{P}_{it}|\bs{X}_{it}])(\theta_0+\sum_{pos=1}^{\#POS-1}\bs{\theta}_{pos}\mathbbm{1}_{\{POS=pos\}}		 + \sum_{tf=1}^{\#TF-1}\bs{\theta}_{tf}\mathbbm{1}_{\{TF=tf\}}) 
	+ \mathtt{log}(\mathbb{E}[{Y}_{it}|\bs{X}_{it}]). 
	\label{eq:estimator}
\end{equation}

The first-stage estimators for $\mathbb{E}[{P}_{it}|\bs{X}_{it}]$ and $\mathbb{E}[{Y}_{it}|\bs{X}_{it}]$ are constructed using a random forest regressor where the number of decision trees to be ensembled are fixed to 100 and use a mean squared error loss function. The features are the same as the ones used in the direct method explained in the previous Section. 
Moreover, to avoid overfitting, we use a cross-fitting approach and split the data into 5-folds with random shuffling. When estimating the second-stage model on each fold, we use the predictions from the first-stage models trained on other folds.

For the second-stage estimator we use the Bayesian Dynamic Generalized Linear Model approach described in Section \ref{sec:inference}. For the model parameters, $\bs{\theta}$, we set the prior mean vector, $\bs{\mu}_{\bs{\theta}} = \bs{0}$, and a diagonal covariance matrix with all variance values set to 10 and the discount factors $\delta^{\bs{\theta}} =1.0$.

The alternative \texttt{Two-stage-PO} approach detailed in Section \ref{sec:semenova} we require three first-stage estimators. For the estimator of $\mathbb{E}[Y_{i}|\bs{X_{i}},P_i]$ we utilize a random forest regressor with Poisson deviance loss, and then construct an estimator for $\mathbb{E}[\mathtt{log}(\mathbb{E}[Y_{i}|\bs{X_{i}},P_i])|\bs{X_{i}}]$ by training another random forest regressor with mean-squared-error loss on $\mathtt{log}$ of predictions of the previous random forest. In addition the estimator for $\mathbb{E}[{P}_{it}|\bs{X}_{it}]$ is constructed in a similar manner to that in the \texttt{Two-stage-PO} approach. In all the random forest regressors, the number of decision trees to be ensembled are fixed to 100. 
The second-stage estimation for the price-sensitivity parameters in the \texttt{Two-stage-PO} approach is performed using a frequentist maximum-likelihood estimation. We use a frequentist approach here since for the score function in \eqref{eq:semenova_score} we need to perform a weighted maximum likelihood estimation using the estimates of $\frac{1}{\mathbb{E}[Y_{i}|\bs{X_{i}},P_i]}$ as weights. Therefore, this weighting scheme cannot be implemented in a meaningful way via Bayesian modeling.

\subsubsection{Comparison of Estimation Performance.}
The final values of estimated price sensitivity parameters $\hat{\alpha} = -\frac{1}{\hat{\theta}}$ after updating over 104 week's worth of data for both direct and two-stage methods are shown in Table \ref{tab:alpha_estimates}. The mean absolute percentage errors (MAPE) between the estimated and true price sensitivity parameters are also shown in Table \ref{tab:alpha_estimates}. We also compute the weighted MAPE (wMAPE) metric where the weights are normalized to 1 based on relative bookings received for various POS/TF combinations i.e., we give higher weight to POS/TF combinations with more bookings. Since the direct approach's performance may change based on the train-validation split and different random initialization of the model, we perform 10 runs and for each run calculate the MAPE and wMAPE and report the means in Table \ref{tab:alpha_estimates}. Note that the individual numbers for the direct approach estimates in Table \ref{tab:alpha_estimates} are from one run that had performance close to the mean (MAPE=23.814\% and wMAPE=0.786\%) and are used as a representative for illustration purposes. 

\begin{table}[!th]
	\centering
	\begin{tabular}{|ccc|cc|cc|cc|}
		\hline
		POS & TF & $\alpha_{\tt{true}} =   -\frac{1}{\theta_{\tt{true}}}$ & $\hat{\alpha}_{\tt{Direct}} =   -\frac{1}{\hat{\theta}}$ & APE & $\hat{\alpha}_{\tt{2-stage-CO}}   = -\frac{1}{\hat{\mu}_{\theta}}$ & APE & $\hat{\alpha}_{\tt{2-stage-PO}}   = -\frac{1}{\hat{\theta}}$ & APE \\ \hline
		0 & 0 & 150 & 182.99 & 22.00\% & 148.82 & 0.79\% & 175.29 & 16.86\% \\ \hline
		0 & 1 & 150 & 197.23 & 31.50\% & 149.85 & 0.10\% & 143.91 & 4.06\% \\ \hline
		0 & 2 & 175 & 192.71 & 10.10\% & 169.99 & 2.86\% & 160.77 & 8.13\% \\ \hline
		0 & 3 & 185 & 191.12 & 3.30\% & 172.07 & 6.99\% & 147.26 & 20.40\% \\ \hline
		0 & 4 & 195 & 200.00 & 2.60\% & 198.75 & 1.93\% & 227.91 & 16.87\% \\ \hline
		0 & 5 & 200 & 183.49 & 8.30\% & 190.38 & 4.81\% & 205.58 & 2.79\% \\ \hline
		0 & 6 & 210 & 207.74 & 1.10\% & 210.28 & 0.13\% & 177.82 & 15.32\% \\ \hline
		0 & 7 & 230 & 220.12 & 4.30\% & 227.54 & 1.07\% & 185.90 & 19.17\% \\ \hline
		0 & 8 & 250 & 207.44 & 17.00\% & 231.98 & 7.21\% & 189.59 & 24.17\% \\ \hline
		0 & 9 & 300 & 226.65 & 24.50\% & 280.52 & 6.49\% & 242.52 & 19.16\% \\ \hline
		1 & 0 & 175 & 264.75 & 51.30\% & 164.18 & 6.18\% & 243.47 & 39.12\% \\ \hline
		1 & 1 & 190 & 295.64 & 55.60\% & 165.44 & 12.93\% & 186.87 & 1.65\% \\ \hline
		1 & 2 & 195 & 285.58 & 46.50\% & 190.34 & 2.39\% & 216.32 & 10.94\% \\ \hline
		1 & 3 & 200 & 282.11 & 41.10\% & 192.94 & 3.53\% & 192.56 & 3.72\% \\ \hline
		1 & 4 & 210 & 301.89 & 43.80\% & 227.14 & 8.16\% & 358.38 & 70.66\% \\ \hline
		1 & 5 & 220 & 265.79 & 20.80\% & 216.28 & 1.69\% & 306.11 & 39.14\% \\ \hline
		1 & 6 & 240 & 319.88 & 33.30\% & 242.32 & 0.97\% & 248.37 & 3.49\% \\ \hline
		1 & 7 & 260 & 350.22 & 34.70\% & 265.54 & 2.13\% & 264.42 & 1.70\% \\ \hline
		1 & 8 & 290 & 319.18 & 10.10\% & 271.60 & 6.35\% & 271.95 & 6.23\% \\ \hline
		1 & 9 & 320 & 367.04 & 14.70\% & 340.60 & 6.44\% & 395.90 & 23.72\% \\ \hline
		&  & \textbf{MAPE} & \textbf{} & \textbf{24.88\%} & \textbf{} & \textbf{4.16\%} & \textbf{} & \textbf{17.36\%} \\ \hline
		&  & \textbf{wMAPE} & \textbf{} & \textbf{0.79\%} & \textbf{} & \textbf{0.17\%} & \textbf{} & \textbf{0.81\%} \\ \hline
	\end{tabular}
	\caption{True price sensitivity parameters used for simulation and their estimates by direct and two-stage methods. Absolute Percentage Error (APE) for direct and two-stage methods estimate of each price-sensitivity parameter, and Mean APE (MAPE) and Weighted MAPE (wMAPE) w.r.t. the true price-sensitivity parameters are shown in the Table. MAPE and wMAPE for the direct method are the average metrics for 10 runs.}
	\label{tab:alpha_estimates}
\end{table}

We see that both the direct approach via wide and deep neural network and the two-stage approaches are able to estimate price-sensitivity parameters in a reasonable range and capture general trends like willingness-to-pay increasing as we get closer to departure and that for a given TF, WTP for POS 1 are higher than that for POS 0. However, the two-stage approaches have lower MAPE errors of $4.16\%$ and $17.36\%$ for \texttt{Two-stage-CO} and \texttt{Two-stage-PO} respectively, as compared to the Wide and Deep approach with an average MAPE error of $24.876 \%$. From the wMAPE metric perspective, the \texttt{Two-stage-CO} approach is substantially better than the \texttt{Two-stage-PO} and the Direct approach which have very similar wMAPE error. These results also show the robustness of the two-stage approach to the minor mis-specification in the linear treatment effect model. 

We note that the \texttt{Two-stage-CO} method provides substantial improvement in the estimates of price sensitivity parameters as compared to other approaches. The estimation for the \texttt{Two-stage-CO} was performed using Bayesian method while that for \texttt{Two-stage-PO} was done via frequentist approach. To rule out the possibility of difference in second stage estimation methodology leading to difference in performance of these two methods, we also performed the estimation for the \texttt{Two-stage-CO} via the same frequentist Poisson GLM approach. We can see from Table \ref{tab:alpha_estimates_bayes_freq} provided in Section \ref{sec:Appendix3} of the Online Appendix that the performance of Bayesian and frequentist estimates for the \texttt{Two-stage-CO} is very close. Therefore, the difference in the performance of these methods cannot be attributed just to the Bayesian approach utilized for the second stage. 

Although the \texttt{Two-stage-CO} and \texttt{Two-stage-PO} approaches performed similarly in simpler simulation setting, in more complex settings the added estimation required in the first-stage for the \texttt{Two-stage-PO} may compound errors and lead to deterioration of estimation performance. These results indicate that the \texttt{Two-stage-CO} is not only computationally efficient but also leads to excellent estimation performance in both simple and complex simulation settings. Therefore, for the numerical study in the next section we choose to perform analysis only on the Direct and the \texttt{Two-stage-CO} approach.

\begin{figure}[h!]
	\centering
	\includegraphics[width=\textwidth]{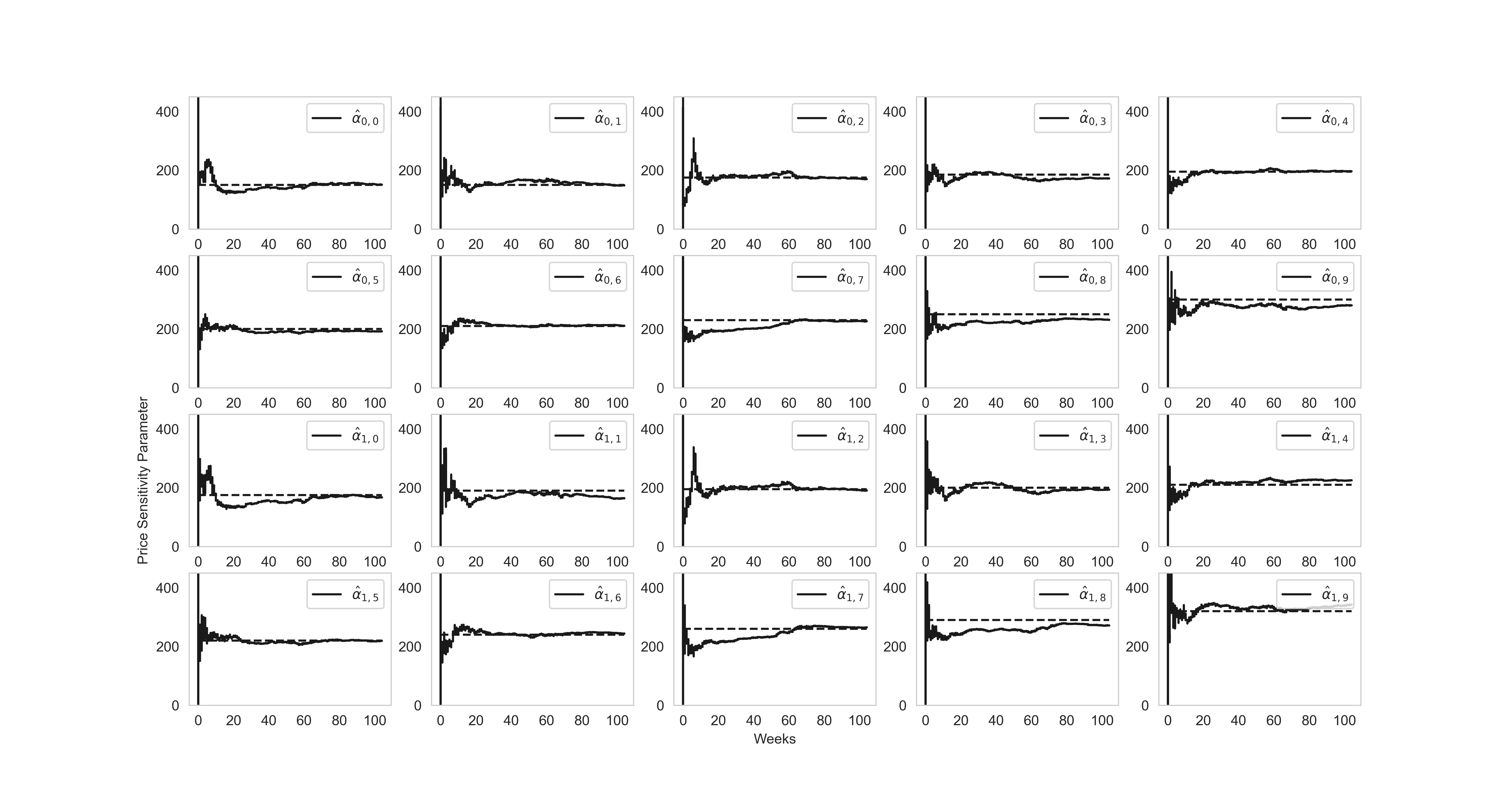}
	\caption{Evolution of price-sensitivity mean estimates ($\hat{\alpha} = \frac{1}{\hat{\mu}_{\theta}}$) over simulation history for the two-stage approach for various (POS, TF) combinations. Dashed line shows the true parameter value.}
	\label{fig:alpha_est}
\end{figure}

The Bayesian DGLM model in the \texttt{Two-stage-CO} approach updates the parameter estimates in an online fashion,  Figure \ref{fig:alpha_est} shows the evolution of the posterior means of price sensitivity parameters ($\hat{\alpha} = \frac{1}{\hat{\mu}_{\theta}}$) for various (POS, TF) combinations. We see that most of the parameters converge to a value close to true parameter value within 20 weeks worth of data. The sparser (POS, TF) combinations e.g., TF 8 take a longer time to converge.

\section{Numerical Study with Real Data}\label{sec:case_study}
In this section, we present results of the proposed methodologies on an anonymized data set comprising of airline sales transactions using an offline estimation study. For real data, since the true price-sensitivity parameters are unknown, it is difficult to validate the quality of the estimators in a manner similar to the simulation study. Instead, we perform a qualitative study to determine the reasonability of the recommended optimal price based on the estimated price-sensitivity parameters by comparing these to the historical prices in the sales transaction data. We also discuss reasonability of trends and patterns observed in the price-sensitivity parameters vis-a-vis expected trends.  

\subsection{Data Description}
The data comprises of historical transactions for a period of 100 weeks for four markets (Origin-Destination pair). The transaction data is for economy compartment and includes daily bookings and daily average price for each departure date for a 180 day booking horizon. 

\subsection{Estimation of Price Sensitivity Parameters}
\subsubsection{Features used for Demand Response Modeling.}
In building the model for price sensitivity estimation, we considered several features including origin and destination, departure time, travel time. Furthermore, departure date related features like yearly seasonality via Fourier series representations based on departure day of year, departure day of week, holidays/special events and booking date related features like seasonality via Fourier series representations based on days before departure were also included. 
For estimating price sensitivity on the real data, we consider price sensitivity to be differentiated by POS, DOW, and Days before Departure (DBD) for each Origin-Destination pair. We use continuous basis functions for representing price sensitivity variation over DBDs. 
\subsubsection{Direct Wide and Deep Approach.}
For direct estimation by the wide and deep method, we use a three layer architecture for the deep part. The features used in the model include the categorical features, and continuous features representing seasonality mentioned above. The wide and deep architecture is trained for each Origin-Destination separately. 
\subsubsection{Two-stage Approach.}
We use neural networks with five layers for the first-stage prediction of price ($\hat{P}_{it}$) and demand ($\hat{Y}_{it}$). The models are trained via a cross-fitting approach, where the flights are split into 10-folds. We observed that by pooling data from other origin-destinations to learn a model for the conditional expected price, the prediction accuracy was improved by 5\% for the four origin-destinations of interest, compared to using data only from one origin-destination.  

For the second-stage estimator we use the Bayesian Dynamic Generalized Linear Model approach described in Section \ref{sec:inference}. For the model parameters, $\bs{\theta}$, we set the prior mean vector, $\bs{\mu}_{\bs{\theta}} = \bs{0}$, and a diagonal covariance matrix with all variance values set to 2 and the discount factors $\delta^{\bs{\theta}} = 0.9995$.

\subsection{Results}
\begin{figure}[h]
\centering
\begin{subfigure}{0.495\textwidth}
	\includegraphics[width=\textwidth]{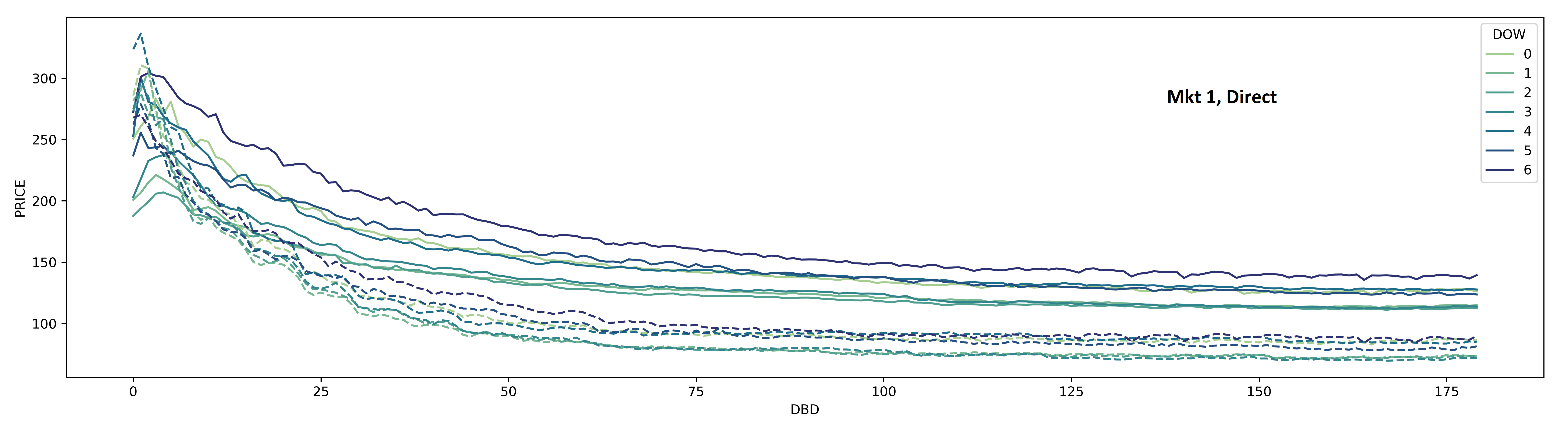}
	\label{fig:price_WD_mkt1}
\end{subfigure}
\begin{subfigure}{0.495\textwidth}
	\includegraphics[width=\textwidth]{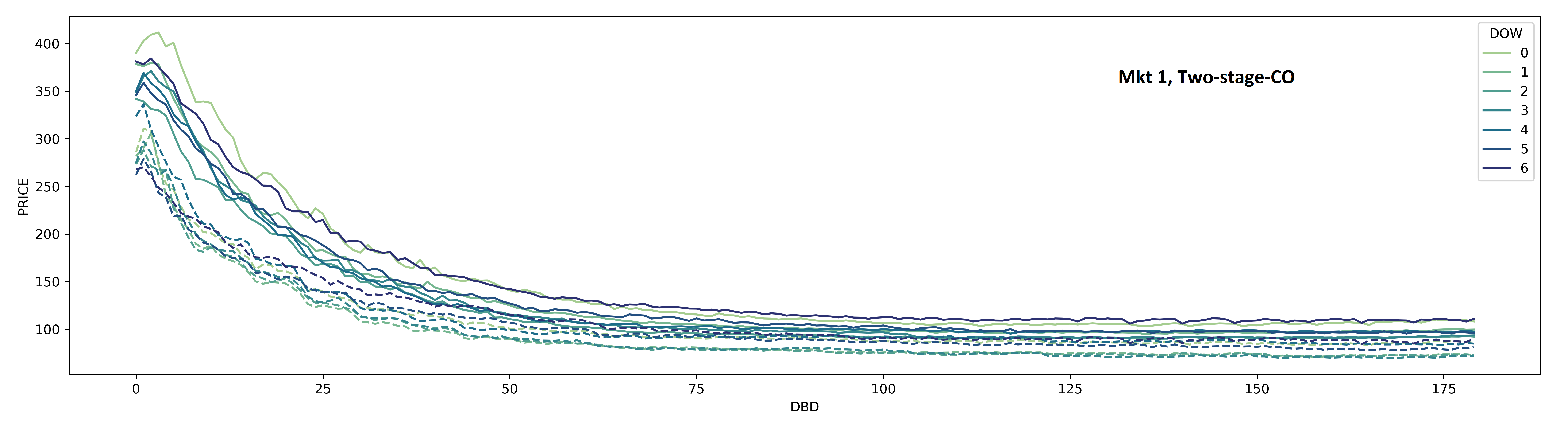}
	\label{fig:price_2S_mkt1}
\end{subfigure}
\begin{subfigure}{0.495\textwidth}
	\includegraphics[width=\textwidth]{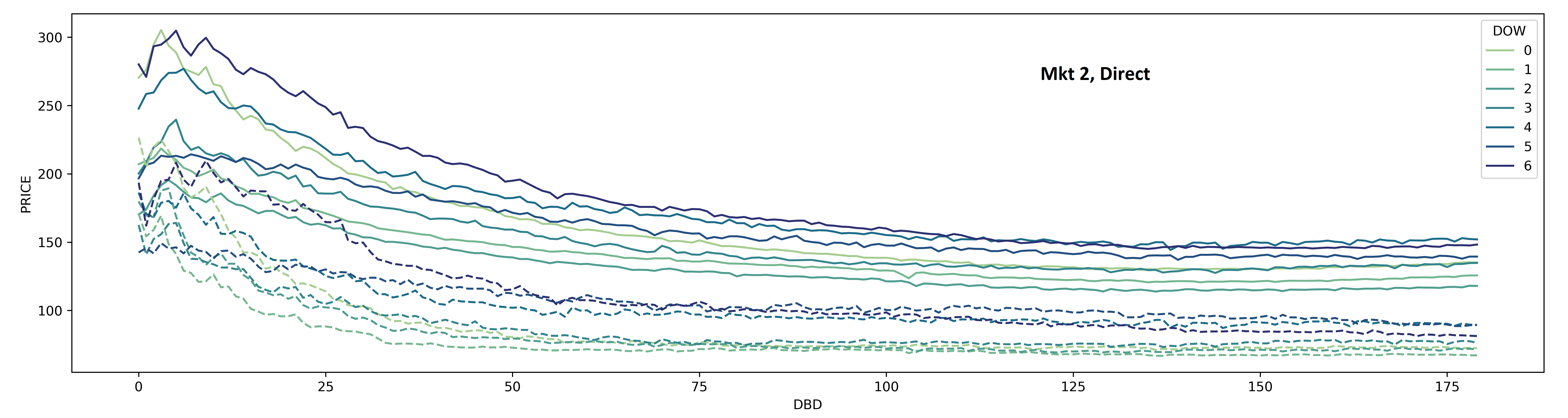}
	\label{fig:price_WD_mkt2}
\end{subfigure}
\begin{subfigure}{0.495\textwidth}
	\includegraphics[width=\textwidth]{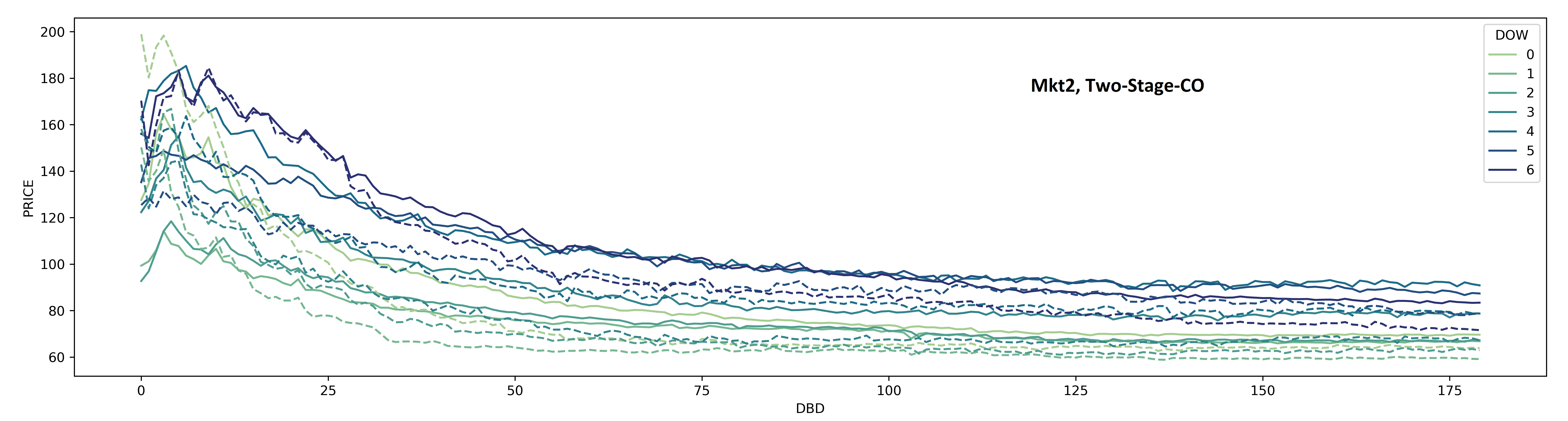}
	\label{fig:price_2S_mkt2}
\end{subfigure}
\begin{subfigure}{0.495\textwidth}
	\includegraphics[width=\textwidth]{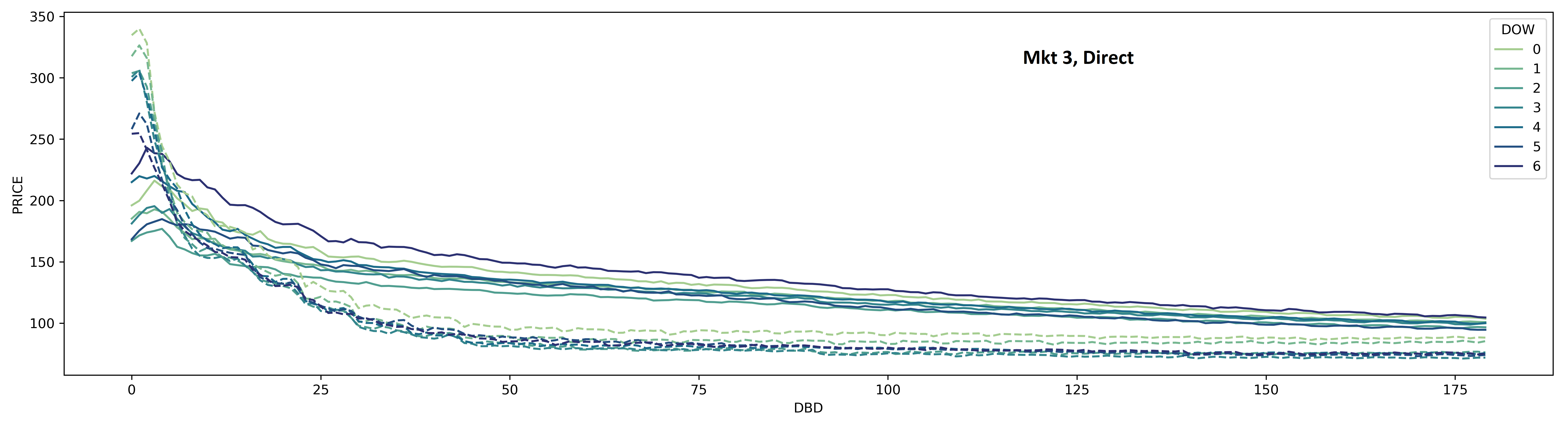}
	\label{fig:price_WD_mkt3}
\end{subfigure}
\begin{subfigure}{0.495\textwidth}
	\includegraphics[width=\textwidth]{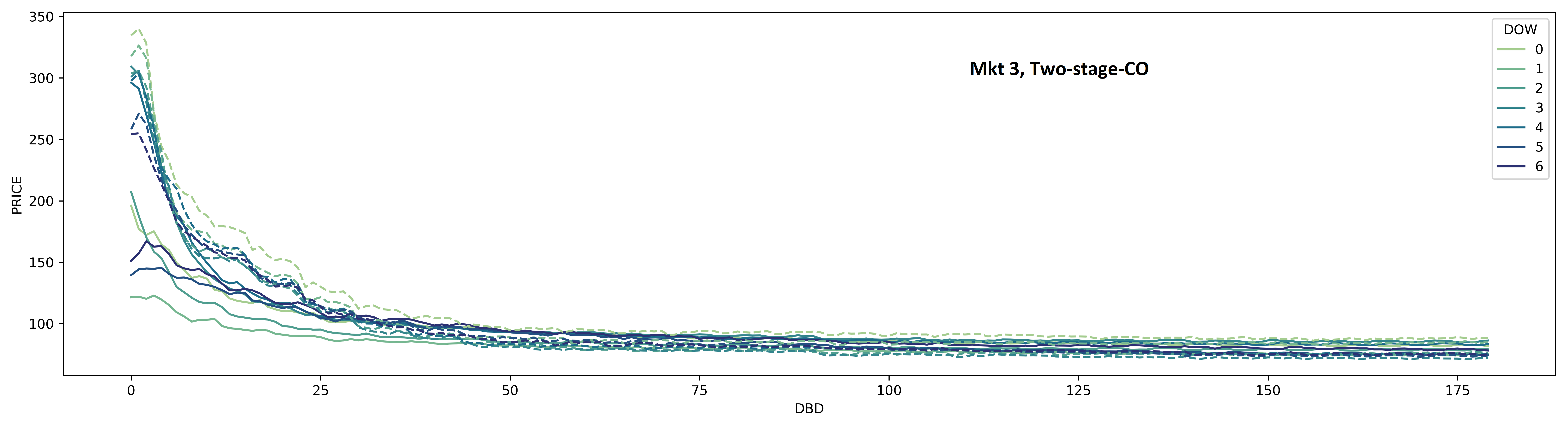}
	\label{fig:price_2S_mkt3}
\end{subfigure}
\begin{subfigure}{0.495\textwidth}
	\includegraphics[width=\textwidth]{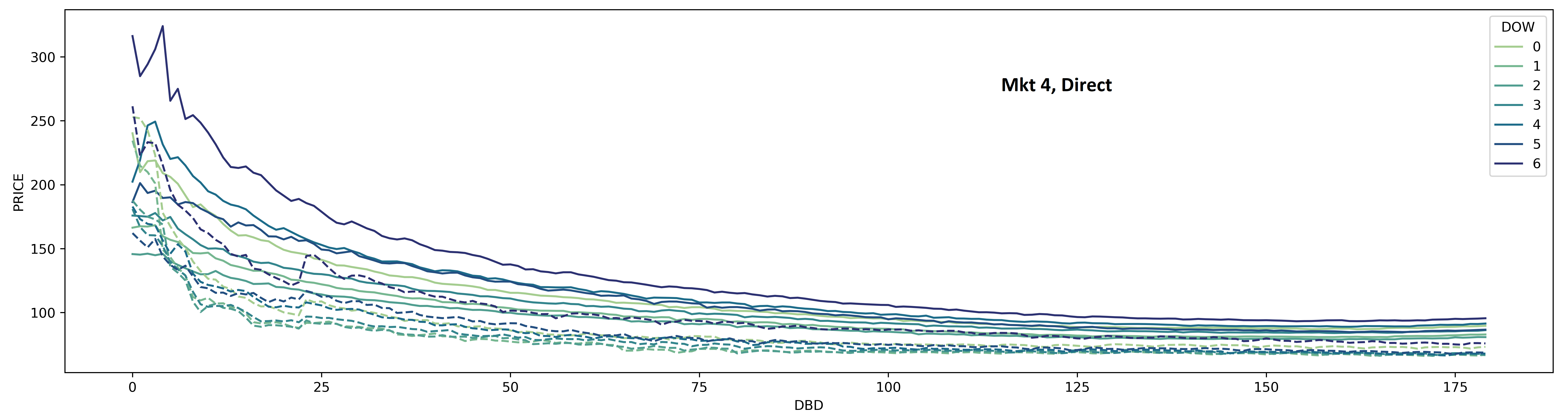}
	\caption{Direct Estimation via Wide and Deep NN}
	\label{fig:price_WD_mkt4}
\end{subfigure}
\begin{subfigure}{0.495\textwidth}
	\includegraphics[width=\textwidth]{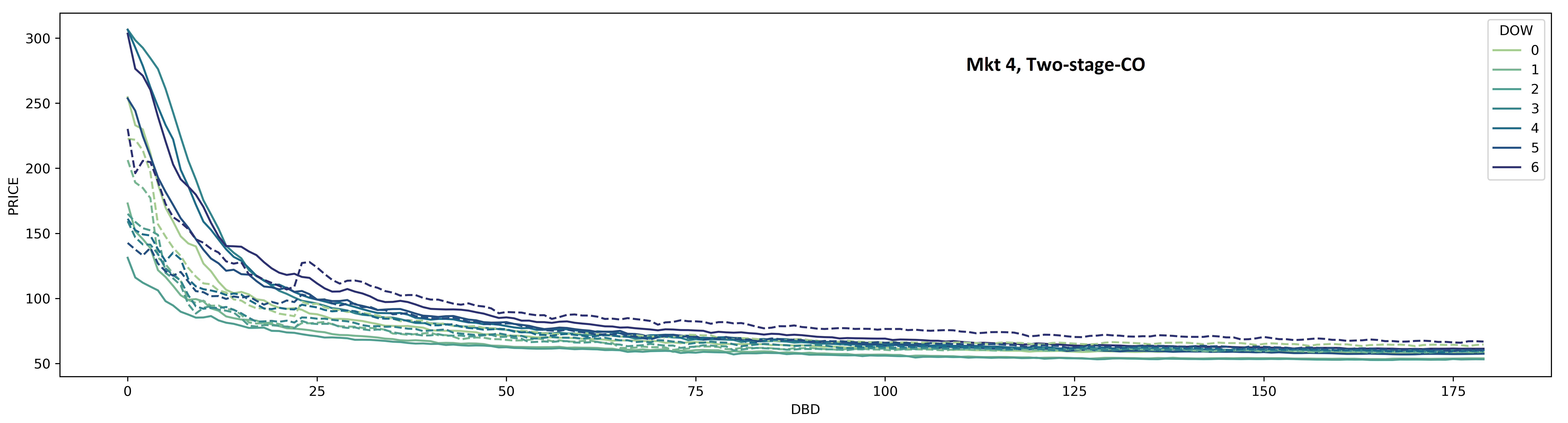}
	\caption{Two-stage-CO Estimation}
	\label{fig:price_2S_mkt4}
\end{subfigure}
\caption{Average optimal price recommended by Direct Wide and Deep method (left panel) and Two-stage-CO method (right panel) for four markets (solid lines). Dotted lines are the average price observed in the transaction data.}
\label{fig:price_recco_real_mkt}
\end{figure}

Unlike the simulation setting, in the real data setting, we cannot compare the estimated price sensitivity parameters to some ground truth. Nevertheless, we can still make some qualitative statements by comparing the price recommended under various estimation schemes and the historical prices. Figure \ref{fig:price_recco_real_mkt} shows the recommended price under Direct and Two-stage estimation approach in solid lines and historical prices in dashed lines over the booking horizon (days before departure) for four airline markets. 

The historical and recommended prices are averaged over other dimensions like departure dates. Note that, if the estimate of the price sensitivity parameter is $\hat{\alpha} = -\frac{1}{\hat{\mu}_{\theta}^T \mathbf{W}}$, then the optimal price given cost $c$ is $c +\hat{\alpha}$. The appropriate cost to use in our setting is the marginal opportunity cost or the bid price computed by the RM system for the inventory position at the time of the booking request. We use the historical marginal opportunity costs stamped in the booking transactions for computing the recommended price corresponding to the price sensitivities estimated with Direct Wide and Deep Neural Network and the \texttt{Two-stage-CO} methods.   

The left panel in Figure \ref{fig:price_recco_real_mkt} shows the evolution of price recommendations based on price-sensitivities estimated using direct estimation approach. We see from these figures, that the recommended prices from the direct approach increase as we go closer to departure on all markets. However, we observe that the recommended prices are much higher than the historical prices for all the markets. Although we do not expect the prices in the historical transactions to be optimal, if a method generates prices considerably higher or lower than historical prices, it is very likely that the price sensitivity estimates are biased.

The right panel in Figure \ref{fig:price_recco_real_mkt} shows the price recommendations based on price-sensitivities estimated using the \texttt{two-stage-CO} approach. We see that similar to the direct estimation approach, the recommended prices are increasing as we go closer to departure. However, unlike the direct estimation approach, the aggressiveness of this increase is much lower and more in line with historical prices. We also see that the price trend across different days of the week (DOW) is similar to that seen in the historical prices. 
\begin{figure}[h!]
\centering
\begin{subfigure}{0.15\textheight}
	\includegraphics[width=\textwidth]{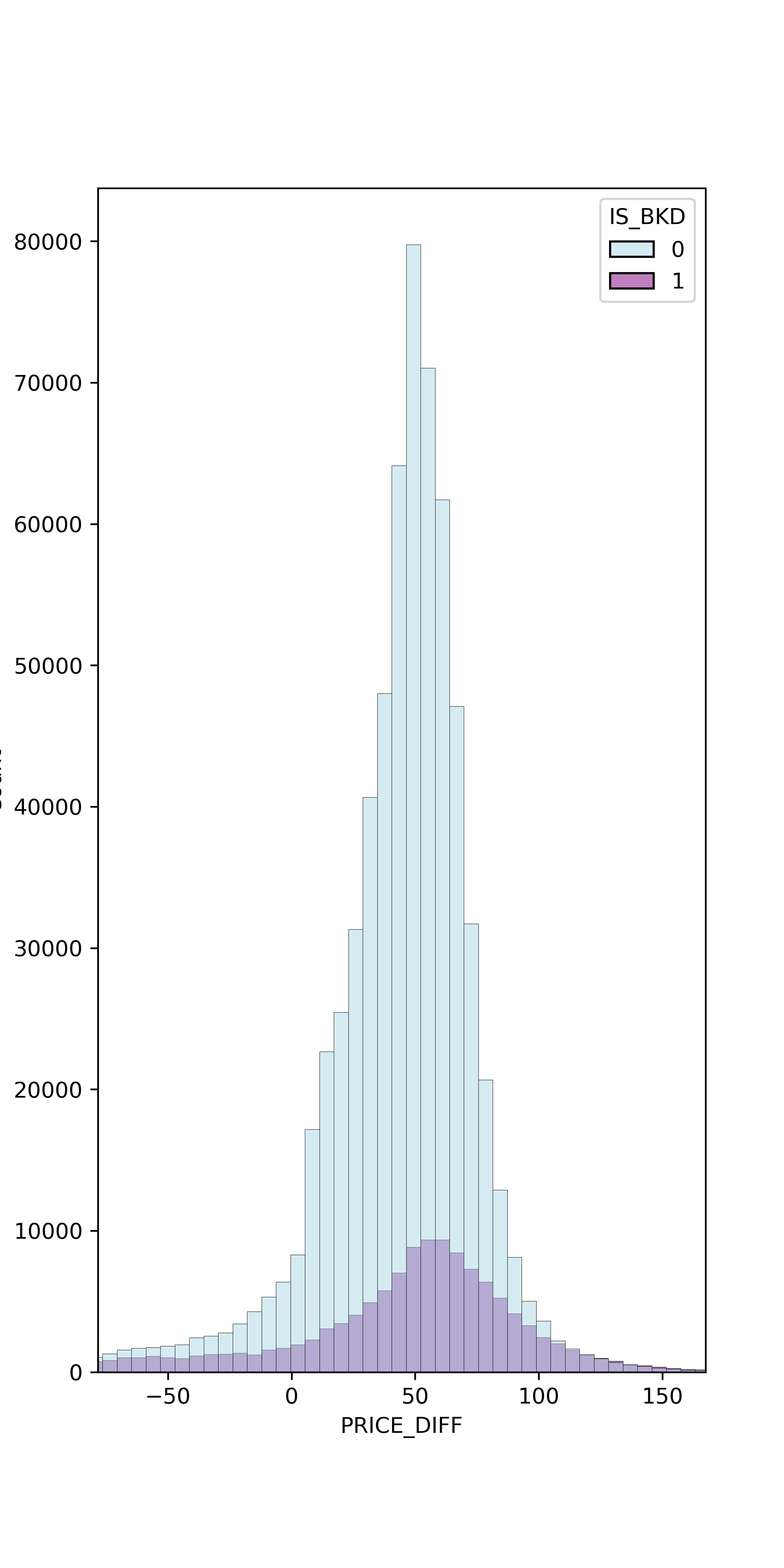}
	\caption{Direct, Mkt 1}
	\label{fig:price_hist_WD_mkt1}
\end{subfigure}
\begin{subfigure}{0.15\textheight}
	\includegraphics[width=\textwidth]{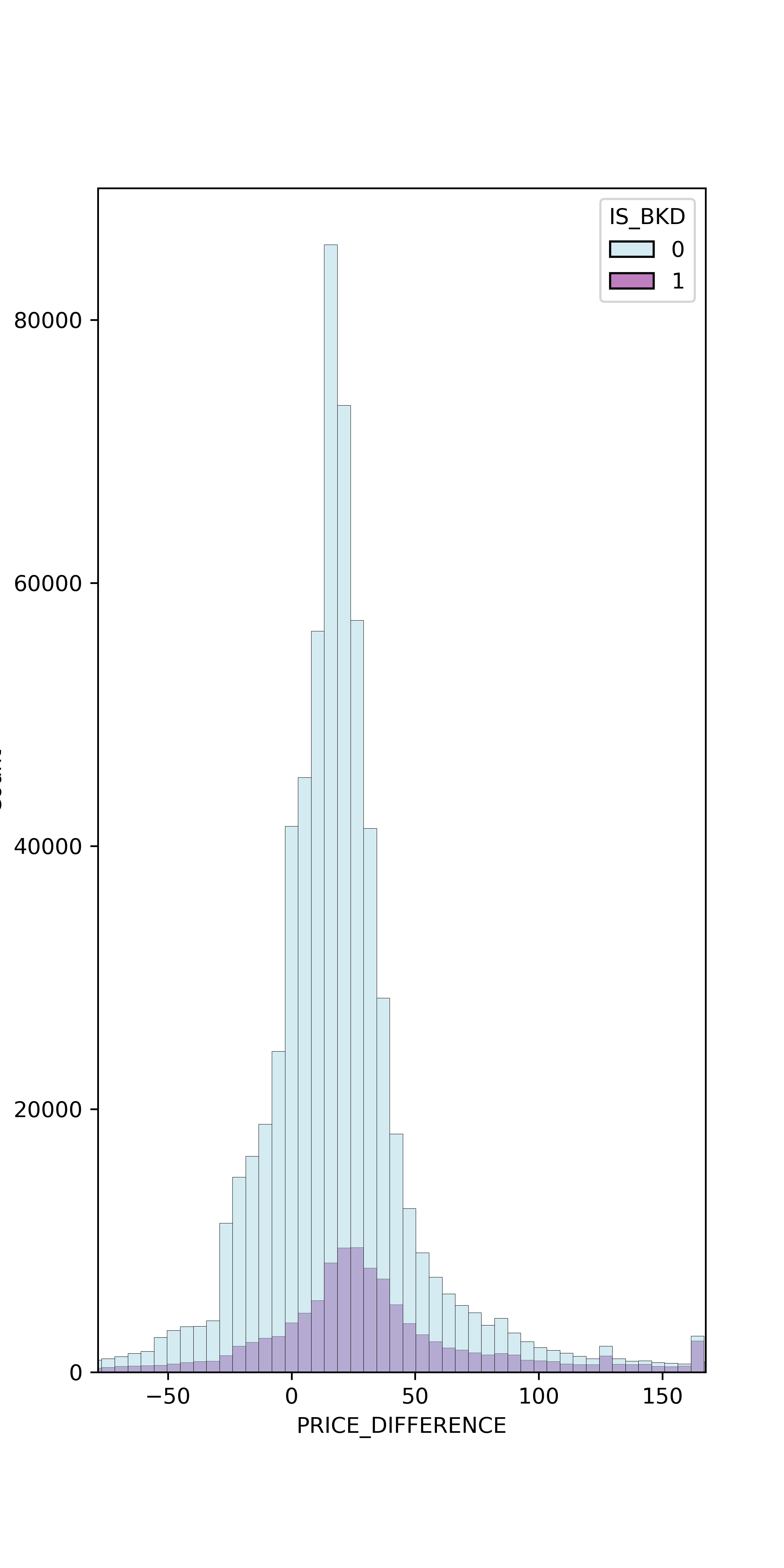}
	\caption{2-stage, Mkt 1}
	\label{fig:price_hist_2S_mkt1}
\end{subfigure}
\begin{subfigure}{0.15\textheight}
	\includegraphics[width=\textwidth]{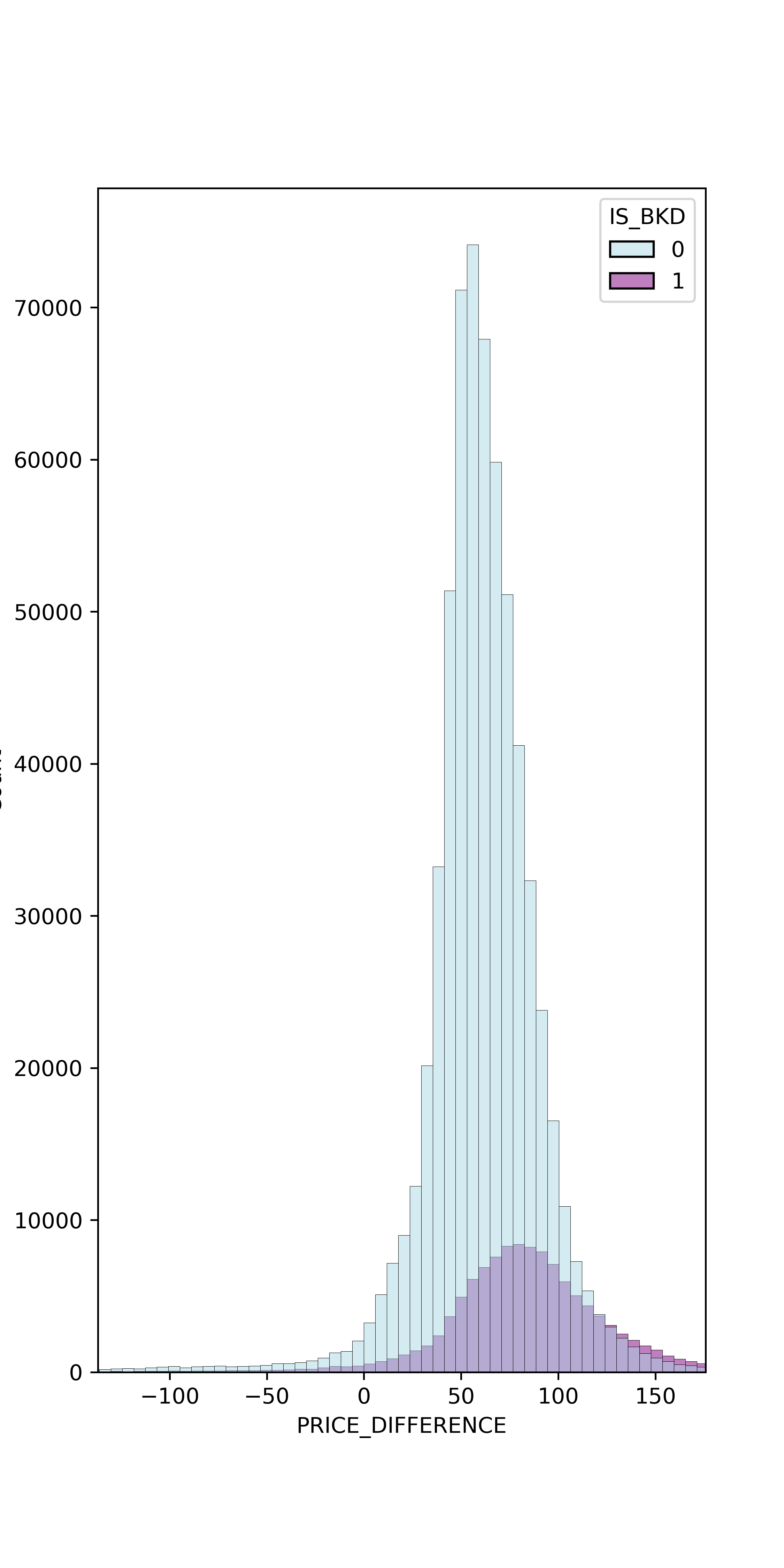}
	\caption{Direct, Mkt 2}
	\label{fig:price_hist_WD_mkt2}
\end{subfigure}
\begin{subfigure}{0.15\textheight}
	\includegraphics[width=\textwidth]{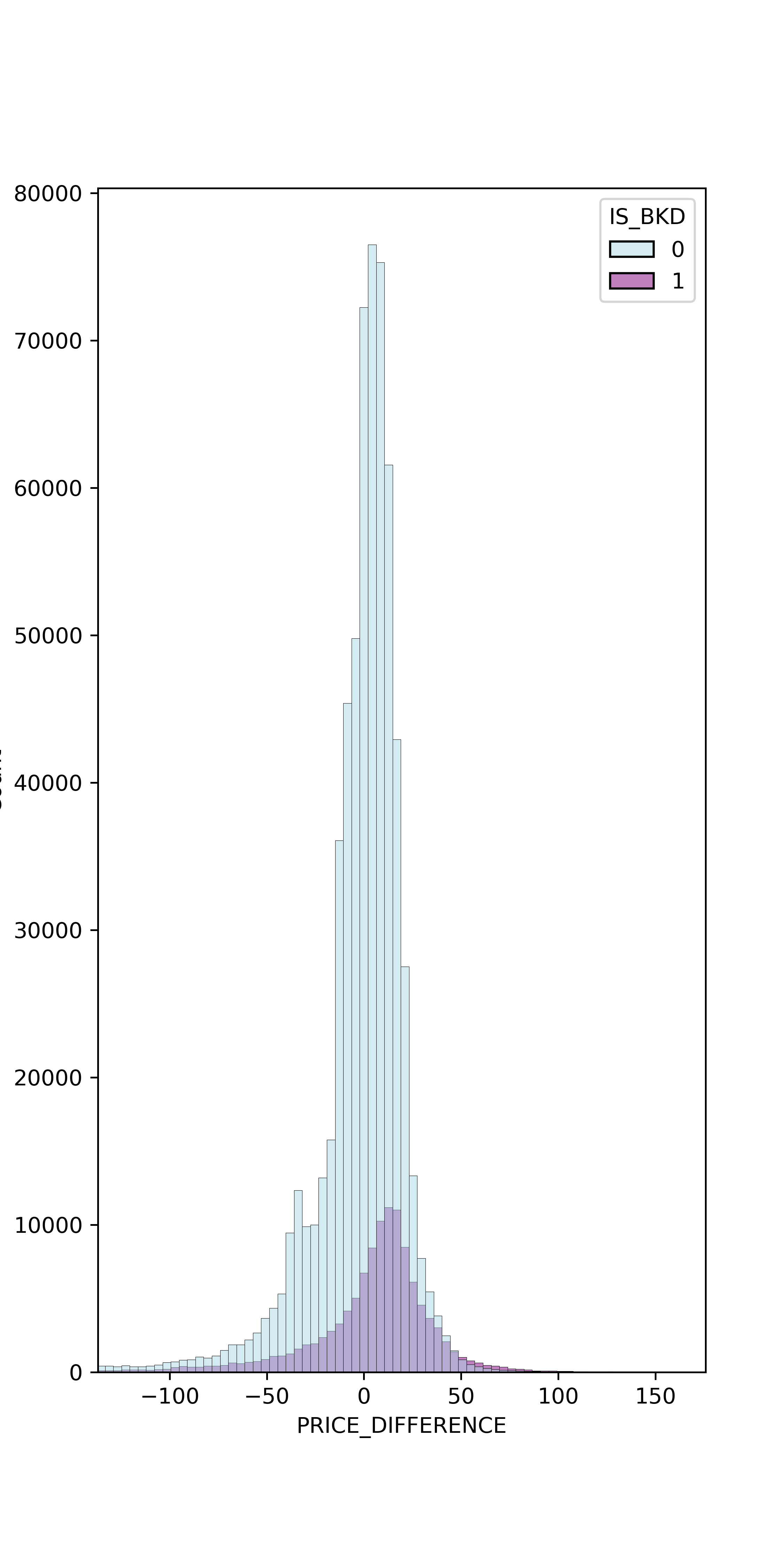}
	\caption{2-stage, Mkt 2}
	\label{fig:price_hist_2S_mkt2}
\end{subfigure}
\begin{subfigure}{0.15\textheight}
	\includegraphics[width=\textwidth]{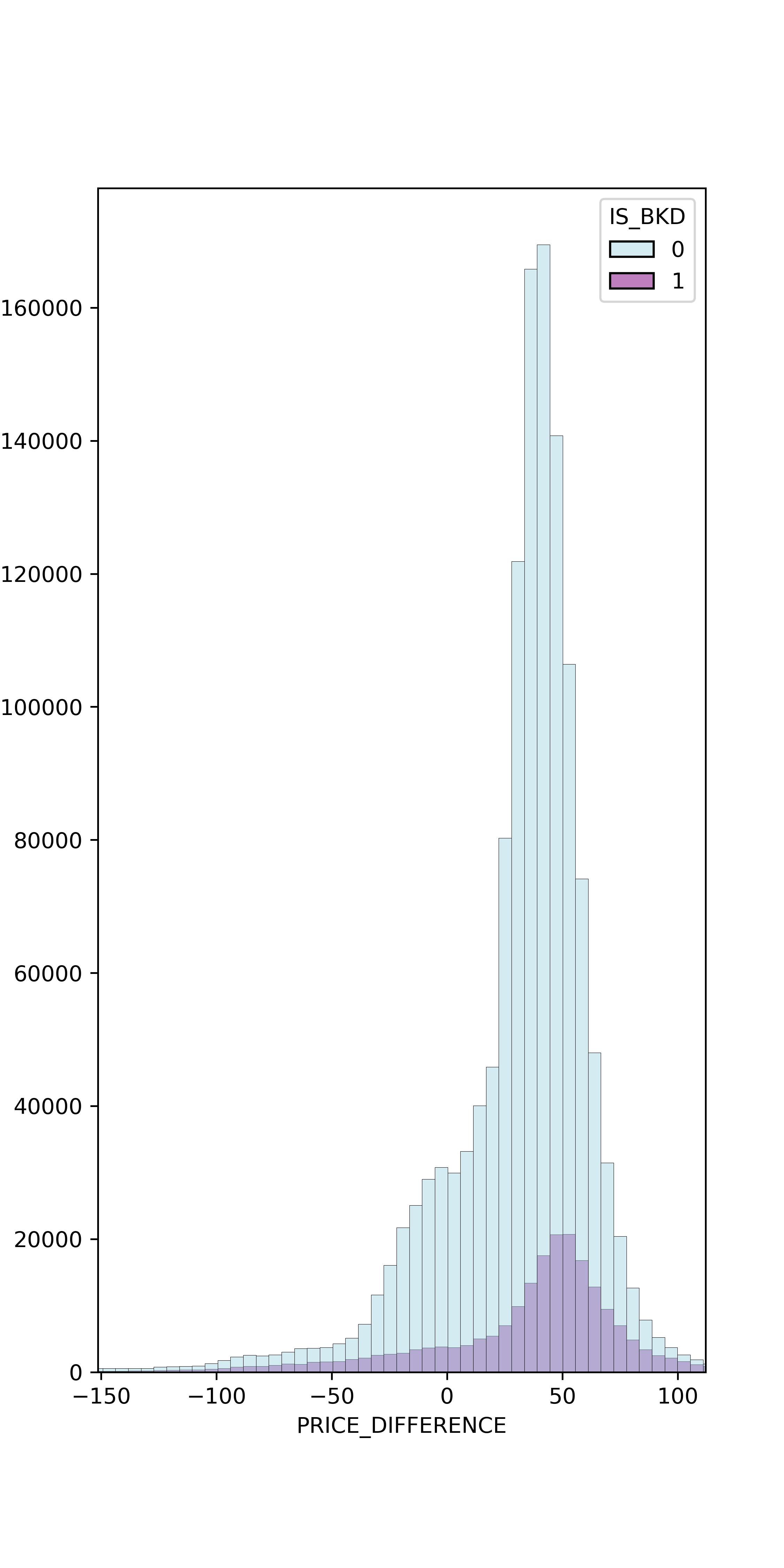}
	\caption{Direct, Mkt 3}
	\label{fig:price_hist_WD_mkt3}
\end{subfigure}
\begin{subfigure}{0.15\textheight}
	\includegraphics[width=\textwidth]{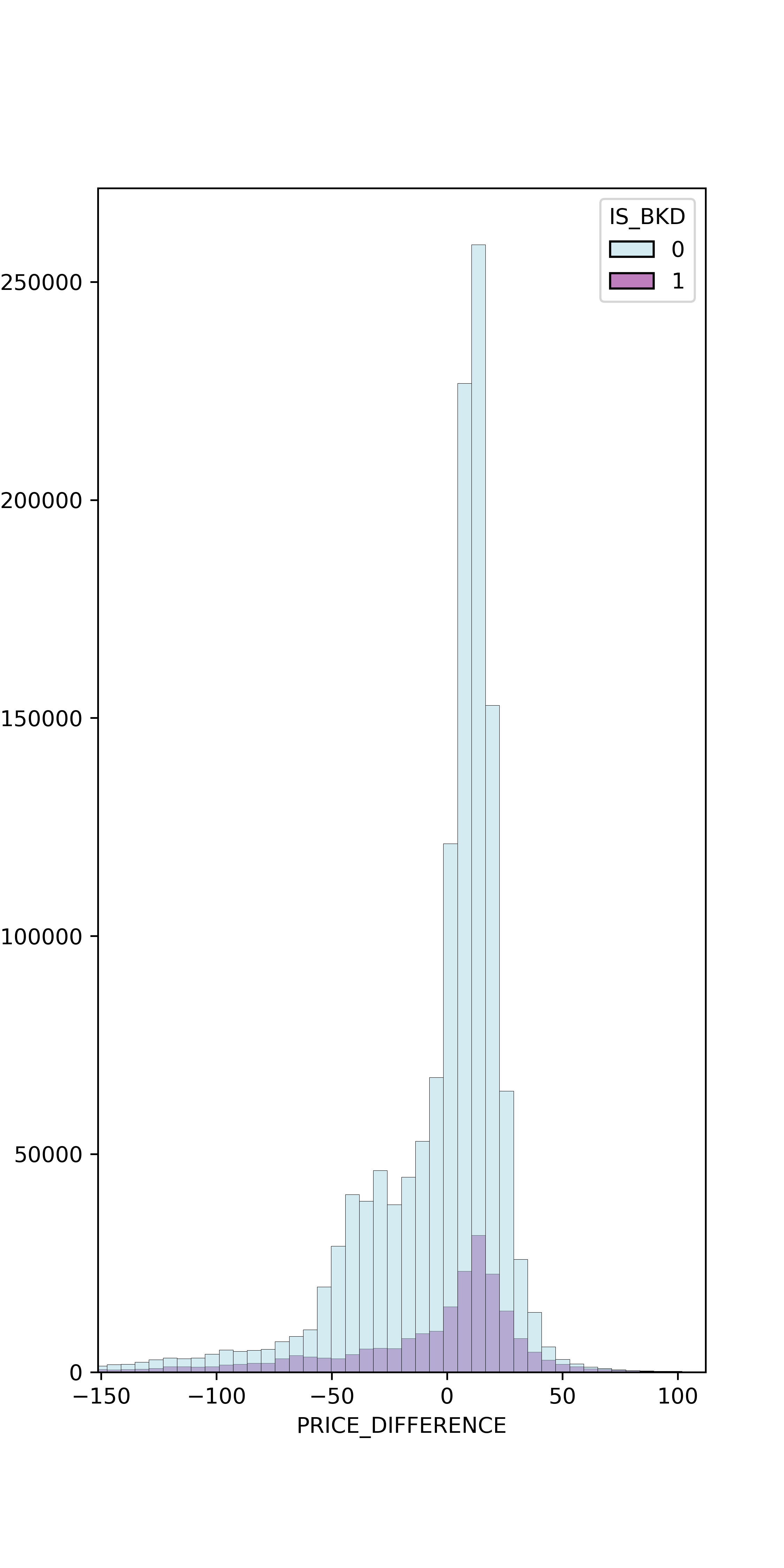}
	\caption{2-stage, Mkt 3}
	\label{fig:price_hist_2S_mkt3}
\end{subfigure}
\begin{subfigure}{0.15\textheight}
	\includegraphics[width=\textwidth]{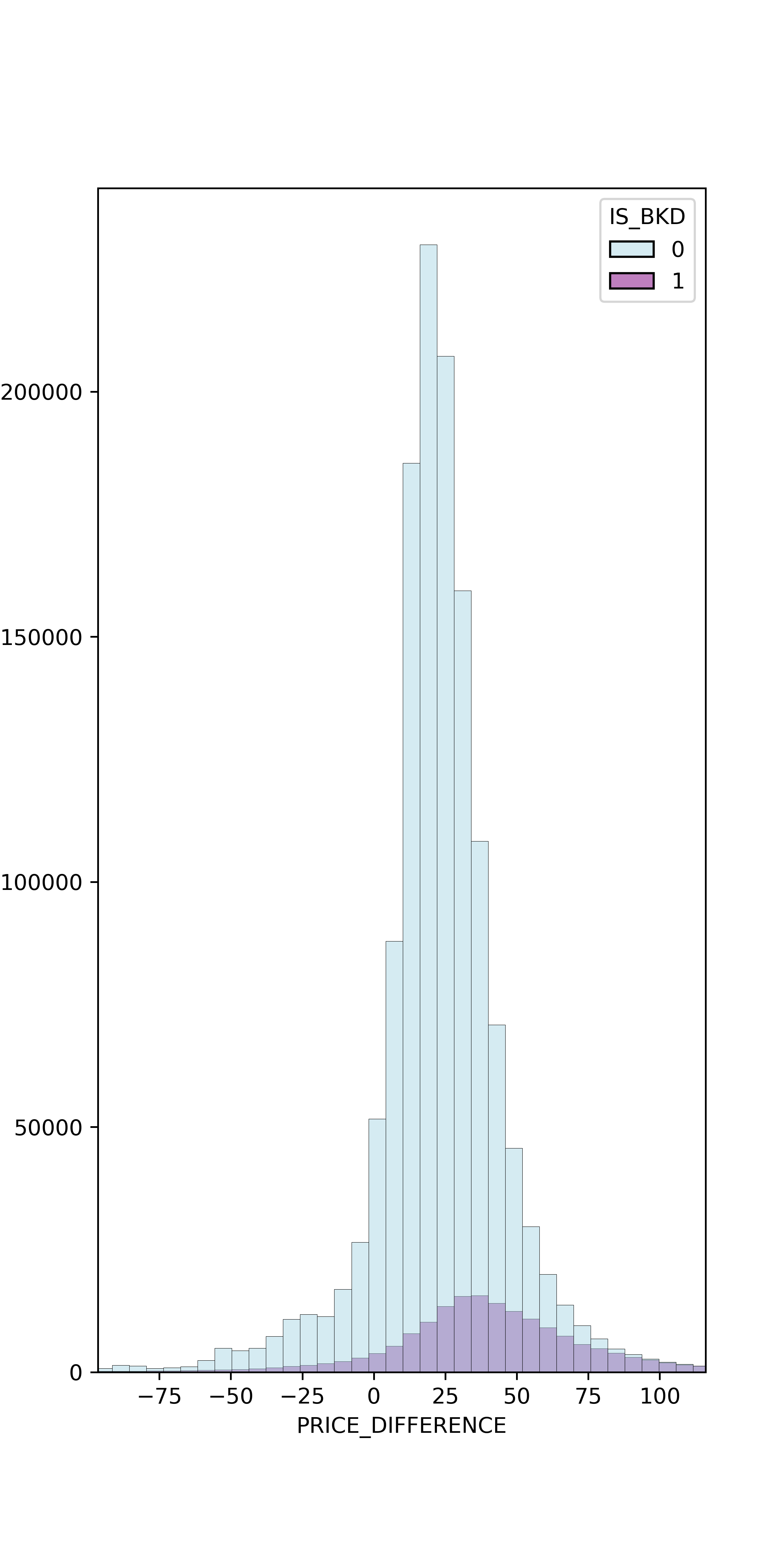}
	\caption{Direct, Mkt 4}
	\label{fig:price_hist_WD_mkt4}
\end{subfigure}
\begin{subfigure}{0.15\textheight}
	\includegraphics[width=\textwidth]{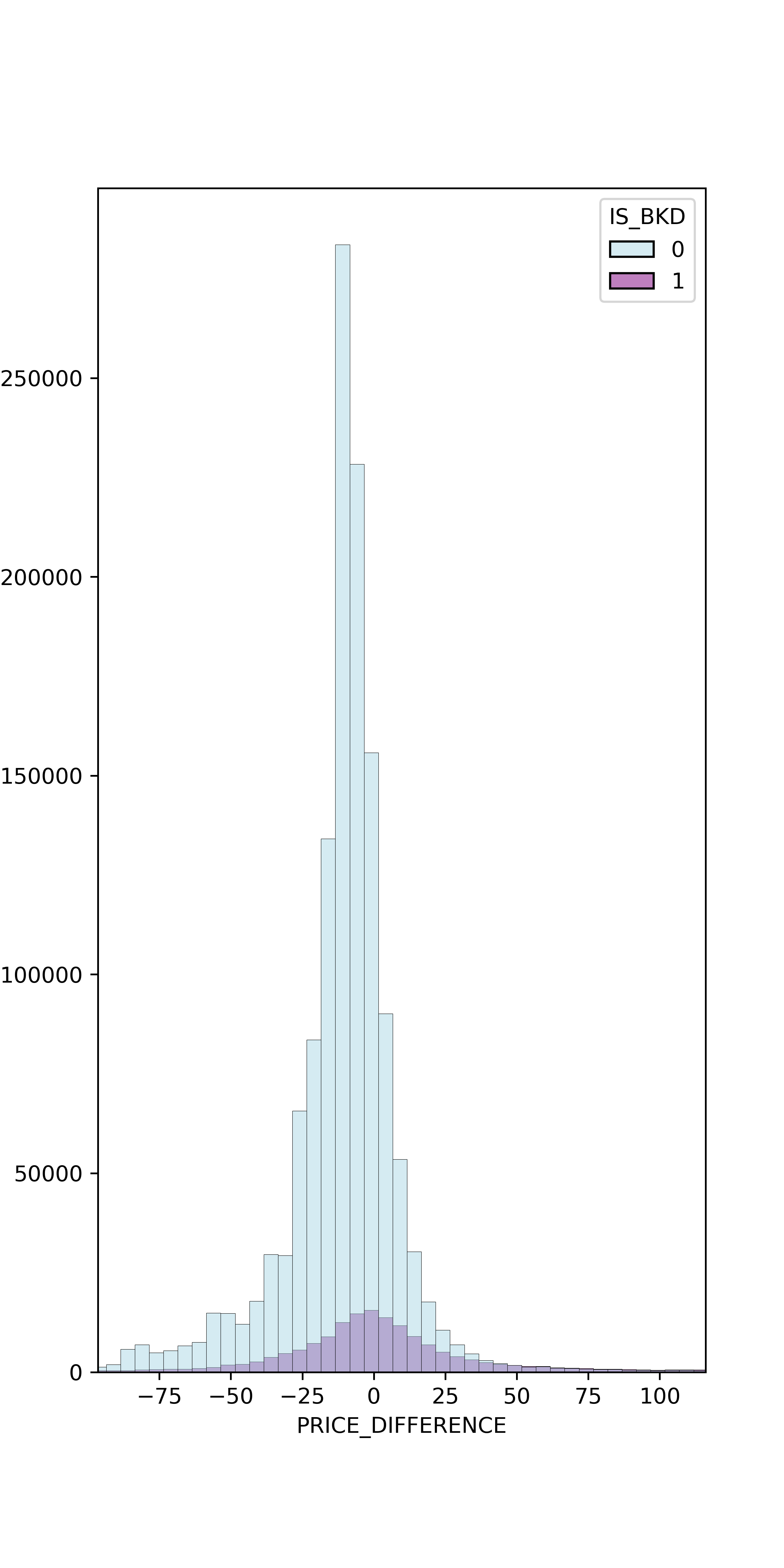}
	\caption{2-stage, Mkt 4}
	\label{fig:price_hist_2S_mkt4}
\end{subfigure}
\caption{Average optimal price recommended by Direct Wide and Deep method (left panel) and Two-stage-CO method (right panel) for four markets (solid lines). Dotted lines are the average price observed in the transaction data.}
\label{fig:price_hist_real_mkt1}
\end{figure}

Figure \ref{fig:price_hist_real_mkt1} shows the distribution of the difference between the recommended and historical prices (positive values indicate that the recommended price is higher than the historical price) for the Direct and \texttt{Two-stage-CO} approach. The distributions are plotted differently corresponding to transactions with positive bookings and those with zero bookings. The prices recommended by the Direct estimation method tend to be much higher than the historical prices (mode occurs at more positive value) for both booking and non-booking related transactions. The prices recommended by the \texttt{Two-stage-CO} approach tend to be closer to historical prices (mode is closer to zero). As with the Direct approach, for most of the markets, the mode of the empirical distribution for the \texttt{Two-stage-CO} method also occurs at slightly positive values indicating that the willingness-to-pay may be higher than that used during the historical period. Although qualitative in nature, these results do provide some evidence that the over-estimation of price-sensitivity parameters observed in the direct estimation approach has been corrected to a large extent by the two-stage estimation approach.

\begin{figure}[h!]
\centering
\includegraphics[width=\textwidth]{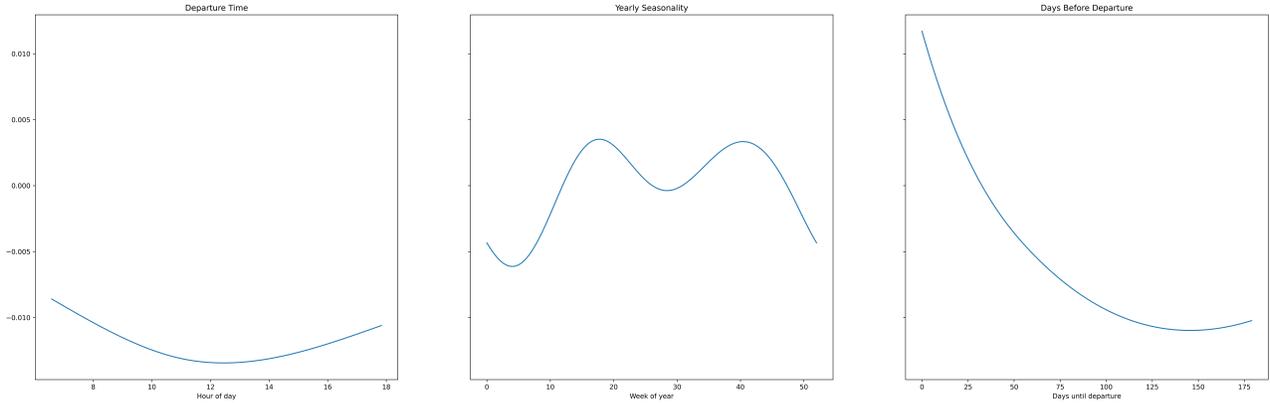}
\caption{Price-sensitivity estimates (${\hat{\mu}_{\theta}}^{factor} \cdot\mathbf{W}^{factor}$) for \texttt{2-stage-CO} associated with various factors for Market 1}
\label{fig:factors_sensitivity}
\end{figure}

As discussed earlier, one of the motivations for keeping the specification of price sensitivity parameters interpretable is that we can easily visualize how these parameters vary across various factors such as departure time, weak of year, days to departure etc. Figure \ref{fig:factors_sensitivity} shows this analysis for Market 1. We can observe the following behaviors for this market:

\begin{itemize}
\item Willingness-to-pay is higher for the morning and evening flight and lower for the flights with departure time during middle of the day. This behavior may indicate significant presence of business travellers.

\item Willingness-to-pay is higher for departure dates in April-May and also from September-October period while it is lowest for the December-February period.

\item For a given departure date, willingness-to-pay increases as we get closer to departure. This is the typical behavior observed in all the markets since leisure customers with higher willingness-to-pay book early and business travelers book later. 
\end{itemize}

These observations are aligned with typical behavior on this market and generates further confidence in the estimates of price-sensitivity parameters from the \texttt{Two-stage-CO} methodology.

\section{Conclusion and Future Research}\label{sec:conclusion}
In this paper, we introduced a Poisson semi-parametric demand response model with feature-dependent price-sensitivity to enable automated dynamic pricing. This model allows one to construct a more flexible model than a fully parametric one. In particular, we allow for the nuisance part to use modern ML techniques while keeping a log-linear specification for the effect of price on demand. After first proposing a direct modeling and  estimation approach for this model via a deep neural network architecture, we then propose a two-stage estimation approach for price-sensitivity estimation based on ideas related to double/debiased ML approach and Neyman orthogonalization of score functions. This approach is advantageous not only from the perspective of practical implementation but also endows robustness to the estimates of price-sensitivity from biases in the estimation of the nuisance part. The second-stage inference of parameters driven by a variational Bayes approach leads to a computationally efficient sequential learning scheme and also enables the implementation of policies for smart price-experimentation based on RL techniques. The numerical studies performed based on simulated data as well a real data demonstrate the superior performance of the two-stage estimation approach as compared to the direct estimation approach.

Although this work was applied to the setting of Airline pricing, we believe that the model and associated estimation techniques discussed in this paper are relevant for retail pricing problem in many industries. Furthermore, the general philosophy of combining traditional fully parametric models with more sophisticated ML techniques e.g., via semi-parametric models, to construct powerful frameworks for decision problems will be helpful beyond the pricing setting discussed in this work. In contrast, for thosesettings where interpretability is not an important concern, extension of the two-stage estimation techniques to fully non-parametric models for dynamic pricing is a promising direction for future research.

\bibliographystyle{IEEEtran}
\bibliography{reference} 

\begin{thebibliography}{10}
\providecommand{\url}[1]{#1}
\csname url@samestyle\endcsname
\providecommand{\newblock}{\relax}
\providecommand{\bibinfo}[2]{#2}
\providecommand{\BIBentrySTDinterwordspacing}{\spaceskip=0pt\relax}
\providecommand{\BIBentryALTinterwordstretchfactor}{4}
\providecommand{\BIBentryALTinterwordspacing}{\spaceskip=\fontdimen2\font plus
\BIBentryALTinterwordstretchfactor\fontdimen3\font minus
  \fontdimen4\font\relax}
\providecommand{\BIBforeignlanguage}[2]{{%
\expandafter\ifx\csname l@#1\endcsname\relax
\typeout{** WARNING: IEEEtran.bst: No hyphenation pattern has been}%
\typeout{** loaded for the language `#1'. Using the pattern for}%
\typeout{** the default language instead.}%
\else
\language=\csname l@#1\endcsname
\fi
#2}}
\providecommand{\BIBdecl}{\relax}
\BIBdecl

\bibitem{hua2021kdd}
J.~Hua, L.~Yan, H.~Xu, and C.~Yang, ``Markdowns in e-commerce fresh retail: A
  counterfactual prediction and multi-period optimization approach,'' in
  \emph{Proceedings of the 27th ACM SIGKDD Conference on Knowledge Discovery \&
  Data Mining}, ser. KDD '21.\hskip 1em plus 0.5em minus 0.4em\relax New York,
  NY, USA: Association for Computing Machinery, 2021, p. 3022–3031.

\bibitem{ye2018kdd}
P.~Ye, J.~Qian, J.~Chen, C.~hung Wu, Y.~Zhou, S.~D. Mars, F.~Yang, and
  L.~Zhang, ``Customized regression model for airbnb dynamic pricing,'' in
  \emph{Proceedings of the 24th ACM SIGKDD International Conference on
  Knowledge Discovery \& Data Mining}, ser. KDD '18.\hskip 1em plus 0.5em minus
  0.4em\relax New York, NY, USA: Association for Computing Machinery, 7 2018,
  p. 932–940.

\bibitem{zou2020counterfactual}
H.~Zou, P.~Cui, B.~Li, Z.~Shen, J.~Ma, H.~Yang, and Y.~He, ``Counterfactual
  prediction for bundle treatment,'' \emph{Advances in Neural Information
  Processing Systems}, vol.~33, 2020.

\bibitem{shalit2017estimating}
U.~Shalit, F.~D. Johansson, and D.~Sontag, ``Estimating individual treatment
  effect: generalization bounds and algorithms,'' in \emph{International
  Conference on Machine Learning}.\hskip 1em plus 0.5em minus 0.4em\relax PMLR,
  2017, pp. 3076--3085.

\bibitem{Athey2019}
S.~Athey and G.~W. Imbens, ``Machine learning methods that economists should
  know about,'' \emph{Annual Review of Economics}, vol.~11, no.~1, pp.
  685--725, 8 2019.

\bibitem{Vinod2021}
B.~Vinod, \emph{The Evolution of Yield Management in the Airline
  Industry}.\hskip 1em plus 0.5em minus 0.4em\relax Springer, 2021.

\bibitem{Angrist2009}
J.~D. Angrist and J.-S. Pischke, \emph{Mostly harmless econometrics}.\hskip 1em
  plus 0.5em minus 0.4em\relax Princeton, NJ: Princeton Univ. Press, 2009.

\bibitem{denBoer2015}
A.~V. {den Boer}, ``Dynamic pricing and learning: Historical origins, current
  research, and new directions,'' \emph{Surveys in Operations Research and
  Management Science}, vol.~20, no.~1, pp. 1--18, 2015.

\bibitem{Cohen2016}
M.~C. Cohen, I.~Lobel, and R.~P. Leme, ``Feature-based dynamic pricing,'' in
  \emph{Proceedings of the 2016 {ACM} Conference on Economics and Computation,
  {EC} '16, Maastricht, The Netherlands, July 24-28, 2016}, V.~Conitzer,
  D.~Bergemann, and Y.~Chen, Eds.\hskip 1em plus 0.5em minus 0.4em\relax {ACM},
  2016, p. 817.

\bibitem{Qiang2016}
S.~Qiang and M.~Bayati, ``Dynamic pricing with demand covariates,'' \emph{arXiv
  e-prints}, p. arXiv:1604.07463, Apr. 2016.

\bibitem{Javanmard2019a}
A.~Javanmard and H.~Nazerzadeh, ``\BIBforeignlanguage{English}{Dynamic pricing
  in high-dimensions},'' \emph{\BIBforeignlanguage{English}{Journal of Machine
  Learning Research (JMLR)}}, vol.~20, p.~49, 2019, id/No 9.

\bibitem{ban_personalized_2021}
G.-Y. Ban and N.~B. Keskin, ``Personalized {Dynamic} {Pricing} with {Machine}
  {Learning}: {High}-{Dimensional} {Features} and {Heterogeneous}
  {Elasticity},'' \emph{Management Science}, vol.~67, no.~9, pp. 5549--5568,
  Sep. 2021, publisher: INFORMS.

\bibitem{Elmachtoub2021}
A.~N. Elmachtoub, V.~Gupta, and M.~L. Hamilton, ``The value of personalized
  pricing,'' \emph{Management Science}, vol.~67, no.~10, pp. 6055--6070, 2021.

\bibitem{Besbes2009}
O.~Besbes and A.~J. Zeevi, ``Dynamic pricing without knowing the demand
  function: Risk bounds and near-optimal algorithms,'' \emph{Operations
  Research}, vol.~57, no.~6, pp. 1407--1420, 2009.

\bibitem{Besbes2015}
O.~Besbes and A.~Zeevi, ``On the (surprising) sufficiency of linear models for
  dynamic pricing with demand learning,'' Hanover, Md, pp. 723--739, 2015.

\bibitem{Perakis2019}
G.~Perakis and D.~Singhvi, ``Dynamic pricing with unknown non-parametric demand
  and limited price changes,'' \emph{Available at SSRN:
  https://ssrn.com/abstract=3336949 or http://dx.doi.org/10.2139/ssrn.3336949},
  2019.

\bibitem{Robinson1988}
P.~M. Robinson, ``Root-n-consistent semiparametric regression,'' p. 931, 1988.

\bibitem{Newey2004}
W.~K. Newey, F.~Hsieh, and J.~M. Robins, ``Twicing kernels and a small bias
  property of semiparametric estimators,'' [Wechselnde Verlagsorte], pp.
  947--962, 2004, literaturverz. S. 961 - 962.

\bibitem{chernozhukov2018double}
V.~Chernozhukov, D.~Chetverikov, M.~Demirer, E.~Duflo, C.~Hansen, W.~Newey, and
  J.~Robins, ``Double/debiased machine learning for treatment and structural
  parameters,'' \emph{The Econometrics Journal}, vol.~21, no.~1, pp. C1--C68,
  2018.

\bibitem{pearl2009causality}
J.~Pearl, \emph{Causality}.\hskip 1em plus 0.5em minus 0.4em\relax Cambridge
  university press, 2009.

\bibitem{newey2003instrumental}
W.~K. Newey and J.~L. Powell, ``Instrumental variable estimation of
  nonparametric models,'' \emph{Econometrica}, vol.~71, no.~5, pp. 1565--1578,
  2003.

\bibitem{hartford2017deep}
J.~Hartford, G.~Lewis, K.~Leyton-Brown, and M.~Taddy, ``Deep iv: A flexible
  approach for counterfactual prediction,'' in \emph{International Conference
  on Machine Learning}.\hskip 1em plus 0.5em minus 0.4em\relax PMLR, 2017, pp.
  1414--1423.

\bibitem{kmenta2010mostly}
J.~Kmenta, ``Mostly harmless econometrics: An empiricist's companion,'' 2010.

\bibitem{angrist1996identification}
J.~D. Angrist, G.~W. Imbens, and D.~B. Rubin, ``Identification of causal
  effects using instrumental variables,'' \emph{Journal of the American
  statistical Association}, vol.~91, no. 434, pp. 444--455, 1996.

\bibitem{darolles2011nonparametric}
S.~Darolles, Y.~Fan, J.-P. Florens, and E.~Renault, ``Nonparametric
  instrumental regression,'' \emph{Econometrica}, vol.~79, no.~5, pp.
  1541--1565, 2011.

\bibitem{chen2012estimation}
X.~Chen and D.~Pouzo, ``Estimation of nonparametric conditional moment models
  with possibly nonsmooth generalized residuals,'' \emph{Econometrica},
  vol.~80, no.~1, pp. 277--321, 2012.

\bibitem{semenova2017estimation}
V.~Semenova, M.~Goldman, V.~Chernozhukov, and M.~Taddy, ``Estimation and
  inference on heterogeneous treatment effects in high-dimensional dynamic
  panels,'' \emph{arXiv preprint arXiv:1712.09988}, 2017.

\bibitem{mackey2018orthogonal}
L.~Mackey, V.~Syrgkanis, and I.~Zadik, ``Orthogonal machine learning: Power and
  limitations,'' in \emph{International Conference on Machine Learning}.\hskip
  1em plus 0.5em minus 0.4em\relax PMLR, 2018, pp. 3375--3383.

\bibitem{oprescu2019orthogonal}
M.~Oprescu, V.~Syrgkanis, and Z.~S. Wu, ``Orthogonal random forest for causal
  inference,'' in \emph{International Conference on Machine Learning}.\hskip
  1em plus 0.5em minus 0.4em\relax PMLR, 2019, pp. 4932--4941.

\bibitem{nekipelov2022regularised}
D.~Nekipelov, V.~Semenova, and V.~Syrgkanis, ``Regularised orthogonal machine
  learning for nonlinear semiparametric models,'' \emph{The Econometrics
  Journal}, vol.~25, no.~1, pp. 233--255, 2022.

\bibitem{athey2019generalized}
S.~Athey, J.~Tibshirani, and S.~Wager, ``Generalized random forests,''
  \emph{The Annals of Statistics}, vol.~47, no.~2, pp. 1148--1178, 2019.

\bibitem{wager2018estimation}
S.~Wager and S.~Athey, ``Estimation and inference of heterogeneous treatment
  effects using random forests,'' \emph{Journal of the American Statistical
  Association}, vol. 113, no. 523, pp. 1228--1242, 2018.

\bibitem{kingma2013auto}
D.~P. Kingma and M.~Welling, ``Auto-encoding variational bayes,'' \emph{arXiv
  preprint arXiv:1312.6114}, 2013.

\bibitem{rezende2014stochastic}
D.~J. Rezende, S.~Mohamed, and D.~Wierstra, ``Stochastic backpropagation and
  approximate inference in deep generative models,'' in \emph{International
  conference on machine learning}.\hskip 1em plus 0.5em minus 0.4em\relax PMLR,
  2014, pp. 1278--1286.

\bibitem{louizos2017causal}
C.~Louizos, U.~Shalit, J.~Mooij, D.~Sontag, R.~Zemel, and M.~Welling, ``Causal
  effect inference with deep latent-variable models,'' in \emph{Proceedings of
  the 31st International Conference on Neural Information Processing Systems},
  2017, pp. 6449--6459.

\bibitem{Sutton2018}
R.~S. {Sutton} and A.~G. {Barto},
  \emph{\BIBforeignlanguage{English}{Reinforcement learning. An
  introduction}}.\hskip 1em plus 0.5em minus 0.4em\relax Cambridge, MA: MIT
  Press, 2018.

\bibitem{Cheng2016}
H.-T. Cheng, L.~Koc, J.~Harmsen, T.~Shaked, T.~Chandra, H.~Aradhye,
  G.~Anderson, G.~Corrado, W.~Chai, M.~Ispir, R.~Anil, Z.~Haque, L.~Hong,
  V.~Jain, X.~Liu, and H.~Shah, ``Wide \& deep learning for recommender
  systems,'' in \emph{Proceedings of the First Workshop on Deep Learning for
  Recommender Systems}, 2016, p. 7–10.

\bibitem{neyman1979c}
J.~Neyman, ``C ($\alpha$) tests and their use,'' \emph{Sankhy{\=a}: The Indian
  Journal of Statistics, Series A}, pp. 1--21, 1979.

\bibitem{west2006bayesian}
M.~West and J.~Harrison, \emph{Bayesian forecasting and dynamic models}.\hskip
  1em plus 0.5em minus 0.4em\relax Springer Science \& Business Media, 2006.

\bibitem{berry2020bayesian}
L.~R. Berry and M.~West, ``Bayesian forecasting of many count-valued time
  series,'' \emph{Journal of Business \& Economic Statistics}, vol.~38, no.~4,
  pp. 872--887, 2020.

\bibitem{goldstein2007bayes}
M.~Goldstein and D.~Wooff, \emph{{B}ayes linear statistics: {T}heory and
  methods}.\hskip 1em plus 0.5em minus 0.4em\relax John Wiley \& Sons, 2007,
  vol. 716.

\bibitem{berry2019bayesian}
L.~R. Berry, ``Bayesian dynamic modeling and forecasting of count time
  series,'' Ph.D. dissertation, Duke University, 2019.

\bibitem{kingma2015adam}
D.~P. Kingma and J.~Ba, ``Adam: A method for stochastic optimization,'' in
  \emph{ICLR}, 2015.

\end{thebibliography}
%
\appendix
\section{Appendix}
	\subsection{Approximation Used for Neyman Orthogonalization}\label{sec:Appendix1}
	Here we discuss the approximation used to derive \eqref{eq:conc_nuisance} in \ref{sec:neyman_derivation} of the main text. 
	
	
	Approximating $\mathtt{exp}(P_{it} \bs{\theta}^T\bs{W_{it}})$ via Taylor series around $\hat{P}_{it} := \mathbb{E}[P_{it}|\bs{X_{it}}]$ we have,
	\begin{eqnarray*}
		\mathbb{E}_{P_{it}|\bs{X_{it}}}[\mathtt{exp}(P_{it} \bs{\theta}^T\bs{W_{it}})] &=& \mathbb{E}_{P_{it}|\bs{X_{it}}}[\mathtt{exp}(\hat{P}_{it}\bs{\theta}^T\bs{W_{it}}) + \bs{\theta}^T\bs{W_{it}}\mathtt{exp}(\hat{P}_{it}\bs{\theta}^T\bs{W_{it}})(P_{it}-\hat{P}_{it}) + O(P_{it}-\hat{P}_{it})^2]\\
		&\approx& \mathtt{exp}(\hat{P}_{it}\bs{\theta}^T\bs{W_{it}}),
	\end{eqnarray*}
	where the approximation is derived by ignoring the second order term if $(\bs{\theta}^T W \sigma_{\epsilon_{it}})^2$, is small.

	\subsection{Details of the Inference in the Second Stage Estimation of Price Sensitivity Parameters}\label{sec:Appendix2}
	
	Noting that for airline booking data, time unit $t$ is relative to the departure date/time of itinerary $i$, the same time units of different itineraries may not be concurrent. For example, $t=0$ for itinerary $i$ corresponds to the time unit farthest from departure and $t=T$ corresponds to the time unit closest to departure. Let the data be sorted by the actual observation time based on $i, t$ indices and the new observation time index be denoted by $j$. The order within the set of $i, t$ indices with the same observation time unit, i.e., the order within concurrent observations from different itineraries, is assigned randomly. For the ease of exposition, in this section we replace $i, t$ with $j$, and define $\bs{H}_j=(P_{it}-\hat{P}_{it}) \bs{W_{it}}$. Also, let $\mathcal{D}_{j-1}$ represent the collection of all observations and features up to and including observation time $j-1$. The dynamic regression model is then defined as:
	\begin{eqnarray}
		\mathtt{log}(\lambda_{j})&=&\bs{\theta}_j^T\bs{H}_j + \mathtt{log}(\hat{Y}_{j}),\nonumber\\\text{where} \quad  \bs{\theta}_j&=&\bs{\theta}_{j-1}+\bs{\upsilon}_j.\label{eq:evolution}
	\end{eqnarray}
	The evolution noise $\bs{\upsilon}_j$ has known mean $\mathbb{E}[\bs{\upsilon}_j|\mathcal{D}_{j-1}]=\bs{0}$ and covariance matrix $\text{Cov}[\bs{\upsilon}_j|\mathcal{D}_{j-1}]=\Upsilon_j$, which is independent from $\bs{\theta}_{j-1}$ given $\mathcal{D}_{j-1}$. Due to the sequential nature of the dynamic regression model, the posterior and estimate of the parameters $\bs{\theta}$ are updated in an online fashion.
	
	Given the posterior moments $\mathbb{E}[\bs{\theta}|\mathcal{D}_{j}]=\bs{\mu}_j$ and $\text{Cov}[\bs{\theta}_j|\mathcal{D}_{j}]={\Sigma}_j$, the prior moments of $\bs{\theta}_{j+1}$ are induced by \eqref{eq:evolution} which in turn leads to prior moments on $\mathtt{log}(\lambda_{j+1})$, $e_{j+1}:=\mathbb{E}[\mathtt{log}(\lambda_{j+1})|\mathcal{D}_{j}]=\bs{H}_{j+1}^{T}\bs{\mu}_j+\mathtt{log}(\hat{Y}_{j+1})$ and $l_{j+1}:=\text{Var}[\mathtt{log}(\lambda_{j+1})|\mathcal{D}_{j}]=\bs{H}_{j+1}^{T}({\Sigma}_j+\Upsilon_{j+1})\bs{H}_{j+1}$. At this stage, a version of variational Bayes concept is applied to update the belief on $\lambda_{j+1}$. The prior distribution for $\lambda_{j+1}$ is assumed to be a Gamma distribution, $\text{Gamma}(a_{j+1},b_{j+1})$, which is the conjugate prior for Poisson. The parameters of the prior are determined either based on moment matching \cite{west2006bayesian} or minimizing forward or backward KL divergence \cite{berry2019bayesian} (for KL one needs to make an additional distributional assumption) via an iterative numerical solver. The posterior for $\lambda_{j+1}$ can then simply be computed as $\text{Gamma}(a_{j+1}+Y_{j+1},b_{j+1}+1)$. The moments of the posterior of $\mathtt{log}(\lambda_{j+1})$ can be computed as $q_{j+1}:=\mathbb{E}[\mathtt{log}(\lambda_{j+1})|\mathcal{D}_{j+1}]=\psi(a_{j+1}+Y_{j+1})-\mathtt{log}(b_{j+1}+1)$ and $\nu_{j+1}:=\text{Var}[\mathtt{log}(\lambda_{j+1})|\mathcal{D}_{j+1}]=\psi'(a_{j+1}+Y_{j+1})$, where $\psi(\cdot)$ and $\psi'(\cdot)$ are the digamma and trigamma functions, respectively. Using linear Bayes updating, the posterior moments $\mathbb{E}[\bs{\theta}_{j+1}|\mathcal{D}_{j+1}]=\bs{\mu}_{j+1}$ and $\text{Cov}[\bs{\theta}_{j+1}|\mathcal{D}_{j+1}]={\Sigma}_{j+1}$ are given by
	\begin{eqnarray}
		\bs{\mu}_{j+1} &=& \bs{\mu}_{j} + ({\Sigma}_j+\Upsilon_{j+1})\bs{H}_{j+1}(q_{j+1}-e_{j+1})/l_{j+1},\nonumber\\
		\Sigma_{j+1} &=& ({\Sigma}_j+\Upsilon_{j+1})-({\Sigma}_j+\Upsilon_{j+1})(\bs{H}_{j+1}\bs{H}_{j+1}^{T})({\Sigma}_j+\Upsilon_{j+1})^T(1-\nu_{j+1}/l_{j+1})/l_{j+1}.\nonumber
	\end{eqnarray}
	
	We specify $\Upsilon_{j+1}$ based on the parsimonious component discounting \cite{west2006bayesian}, where $\Upsilon_{j+1}=\Sigma_j(1-\delta)/\delta$ and $\delta$ is typically in [0.9,1.0].
	
	Although the above procedure is efficient, the variational step to find the conjugate distributions's parameters still requires an iterative numerical computation. To circumvent the need to use an iterative solver we propose to take a Laplace approximation instead and derive a closed-form update using the principal branch of Lambert W function (inverse of $\omega \mapsto \omega \mathtt{exp}(\omega)$). Having $e_{j+1}$ and $l_{j+1}$ and assuming a normal distribution for $P(\mathtt{log}(\lambda_{j+1}))|\mathcal{D}_{j}) \sim \mathcal{N}(e_{j+1},l_{j+1})$, $q_{j+1}=\arg\!\max_{\omega}\mathtt{log}\big(\text{Pois}(Y_{j+1};\mathtt{exp}(\omega))\mathcal{N}(\omega;e_{j+1},l_{j+1})\big)$ and $\nu_{j+1}$ is the inverse of the curvature of negative log-posterior at $q_{j+1}$:
	\begin{eqnarray}
		q_{j+1}&=&\mathtt{log}\Big(\text{LambertW}\big(l_{j+1}\mathtt{exp}(Y_{j+1}l_{j+1}+e_{j+1})\big)/l_{j+1}\Big),\nonumber\\
		\nu_{j+1}&=&l_{j+1}/(1+l_{j+1}\mathtt{exp}(q_{j+1})).\nonumber
	\end{eqnarray}
	
	\subsection{Constructing Bayes Greedy and RL-based Pricing Policies}\label{sec:price-opt}
	
	\subsubsection{Bayes Greedy Policy}\label{sec: bayes_greedy}
	
	As a second option to compute \eqref{eq:bay-rew_} of the main text, we can approximate the expectation in \eqref{eq:bay-rew_} without any specific distributional assumption by a second order Taylor expansion of the exponential term around the mean of parameters:
	\begin{equation}
		\begin{split}
			&\mathbb{E}_{\pi^{*}(\bs{\theta})}[ \mathtt{exp}(p\bs{\theta}^T\bs{W}_{it})(p-c)] \approx \\&(p-c) \mathbb{E}_{\pi^{*}(\bs{\theta})}[ \mathtt{exp}(p\bs{\mu}_{\theta}^T\bs{W}_{it}) + \mathtt{exp}(p\bs{\mu}_{\theta}^T\bs{W}_{it})(\bs{\theta}-\bs{\mu}_{\theta})^T(p\bs{W}_{it})+\frac{1}{2}(\bs{\theta}-\bs{\mu}_{\theta})^T(p\bs{W}_{it})(p\bs{W}_{it})^T(\bs{\theta}-\bs{\mu}_{\theta})]\\
			&=(p-c)\mathtt{exp}(p\bs{\mu}_{\theta}^T\bs{W}_{it})\left(1+\frac{1}{2}p^2\bs{W}_{it}^T\Sigma_{\theta}\bs{W}_{it}\right).\label{eq:taylor-eq_}
		\end{split}
	\end{equation}
	
	Assuming $\pi^{*}(\bs{\theta})$ is a multivariate normal distribution leads us to a third approach to approximate \eqref{eq:bay-rew_}. Under this assumption and considering the random variable $p\bs{\theta}^T\bs{W}_{it}$ should lie within $(-\infty,0)$, we can derive a closed-form for \eqref{eq:bay-rew_} and calculate the expectation in \eqref{eq:bay-rew_} as follows:
	\begin{equation}
		\begin{split}
			\mathbb{E}_{\pi^{*}(\bs{\theta})}[ \mathtt{exp}(p\bs{\theta}^T\bs{W}_{it})(p-c)]&=(p-c)\int_{-\infty}^{0}\frac{\mathtt{exp}(z)}{Z\left(\frac{-p\bs{\mu}_{\theta}^T\bs{W}_{it}}{p\sqrt{\bs{W}_{it}^T\Sigma_{\theta}\bs{W}_{it}}}\right)}\frac{\zeta\left(\frac{z-p\bs{\mu}_{\theta}^T\bs{W}_{it}}{p\sqrt{\bs{W}_{it}^T\Sigma_{\theta}\bs{W}_{it}}}\right)}{p\sqrt{\bs{W}_{it}^T\Sigma_{\theta}\bs{W}_{it}}}=\\
			&(p-c)\mathtt{exp}\left(p\bs{\mu}_{\theta}^T\bs{W}_{it}+\frac{p^2\bs{W}_{it}^T\Sigma_{\theta}\bs{W}_{it}}{2}\right)\frac{Z\left(\frac{-\bs{\mu}_{\theta}^T\bs{W}_{it}-p\bs{W}_{it}^T\Sigma_{\theta}\bs{W}_{it}}{\sqrt{\bs{W}_{it}^T\Sigma_{\theta}\bs{W}_{it}}}\right)}{Z\left(\frac{-\bs{\mu_{\theta}}^T\bs{W}_{it}}{\sqrt{\bs{W}_{it}^T\Sigma_{\theta}\bs{W}_{it}}}\right)},
		\end{split}\label{eq:tn-eq_}
	\end{equation}
	where $\zeta(\cdot)$ and $Z(\cdot)$ denote the standard normal PDF and CDF, respectively. In contrast to the plug-in estimator in \eqref{eq:plugin_}, both \eqref{eq:tn-eq_} and \eqref{eq:taylor-eq_} depend on $\Sigma_{\theta}$ in addition to the estimated mean of $\bs{\theta}$. For real-time pricing by using the utility in \eqref{eq:tn-eq_} or \eqref{eq:taylor-eq_}, we can discretize $[p^{lb}_{it},p^{ub}_{it}]$ and query the approximated utility to find the price maximizing the expected margin contribution.
	
	\subsubsection{Pricing with Learning}\label{sec: bayes_learn}
	The estimation of price sensitivities via the Bayesian DGLM approach allows one to leverage ideas from Reinforcement Learning for developing dynamic price experimentation policies based on methods such as Upper Confidence Bound (UCB) or Thompson sampling (TS) \cite{Sutton2018}. Such policies incorporate the knowledge about current uncertainty in the price sensitivity parameters to judiciously randomize prices and typically converge to the optimal price faster than the Bayes greedy policies. For TS, the policy is to draw samples from the posterior so as to select a price with probability equal to the posterior probability that the price results in the highest margin contribution. With the normal distributional assumption for the posterior of the parameters, implementing TS is rather straightforward, where a sample ($\bar{\bs{\theta}}$) from the posterior of the parameters is taken (with rejection if $\bar{\bs{\theta}}^T\bs{W}_{it}\leq0$) and the price that maximizes $\mathtt{exp}(p\bar{\bs{\theta}}^T\bs{W}_{it})(p-c)$ is selected. The offered price under TS is $\min\{\max\{p^{lb}_{it}, c-1/\bar{\bs{\theta}}^T\bs{W}_{it}\}, p^{ub}_{it}\}$, which is similar to \eqref{eq:plug-sol_} in the main text with $\bs{\mu}_{\theta}$ replaced by the parameter sample.
	
	For UCB, the policy is to select the price that maximizes a desired quantile of the margin contribution (the choice of quantile can be time dependent). Since the Bayesian DGLM approach doesn't enforce a distributional assumption for $\pi^{*}(\bs{\theta})$, we will assume a normal distributional approximation to develop the pricing policy under UCB. The UCB in this case is approximated by $\mathbb{E}[\mathtt{exp}(p\bs{\theta}^T\bs{W}_{it})(p-c)]+\xi\text{Std}[\mathtt{exp}(p\bs{\theta}^T\bs{W}_{it})(p-c)]$, where $\text{Std}[\cdot]$ represents standard deviation, $\xi=\sqrt{2}\mathtt{erf}^{-1}(2\alpha-1)$, and $\alpha$ denotes the desired quantile. By first-order Taylor expansion we can approximate the UCB as
	\begin{equation}
		\begin{split}
			\text{UCB}(\alpha) &\approx \mathbb{E}[\mathtt{exp}(p\bs{\theta}^T\bs{W}_{it})(p-c)]+\xi\text{Std}[\mathtt{exp}(p\bs{\theta}^T\bs{W}_{it})(p-c)] \\&\approx (p-c)\mathtt{exp}(p\bs{\mu}_{\theta}^T\bs{W}_{it})+\xi(p-c)\mathtt{exp}(p\bs{\mu}_{\theta}^T\bs{W}_{it})p\sqrt{\bs{W}_{it}^T\Sigma_{\theta}\bs{W}_{it}}.\label{eq:ucbtaylor-eq_}
		\end{split}
	\end{equation}
	Since the approximated UCB in \eqref{eq:ucbtaylor-eq_} is differentiable and log-concave for $p^{lb}_{it}>c$, the price maximizing \eqref{eq:ucbtaylor-eq_} is either the root of its gradient (root of a quadratic polynomial which can be easily calculated) or one of the price boundaries, $p^{lb}_{it}$ or $p^{ub}_{it}$.
	
	By further assuming that the posterior distribution of the parameters is multivariate normal, we can derive a closed-form for UCB. Specifically, we can compute the desired quantile of the margin contribution based on standard normal quantile (inverse CDF) function. Under this assumption and considering that $p\bs{\theta}^T\bs{W}_{it}$ lies within $(-\infty,0)$, the quantile function of the margin contribution is equal to
	\begin{equation}
		\text{UCB}(\alpha)= \mathtt{exp}\left(p\bs{\mu}_{\theta}^T\bs{W}_{it} + p\sqrt{\bs{W}_{it}^T\Sigma_{\theta}\bs{W}_{it}}Z^{-1}\left(\alpha Z\left(\frac{-\bs{\mu}_{\theta}^T\bs{W}_{it}}{\sqrt{\bs{W}_{it}^T\Sigma_{\theta}\bs{W}_{it}}}\right)\right)\right) (p-c).\label{eq:ucbtn-eq_}
	\end{equation}
	Then the optimal price with respect to the UCB utility function in \eqref{eq:ucbtn-eq_} has the following closed-form solution
	\begin{equation}
		P_{it}^{\ast} = \min\left\{\max\left\{p^{lb}_{it}, c-1/\left(\bs{\mu}_{\theta}^T\bs{W}_{it} +\sqrt{\bs{W}_{it}^T\Sigma_{\theta}\bs{W}_{it}}Z^{-1}\left(\alpha Z\left(\frac{-\bs{\mu}_{\theta}^T\bs{W}_{it}}{\sqrt{\bs{W}_{it}^T\Sigma_{\theta}\bs{W}_{it}}}\right)\right)\right)\right\}, p^{ub}_{it}\right\}.\label{eq:ucbtn-sol_}
	\end{equation}

	\subsection{Additional Details about the Simulation Setup Used in the Airline Setting}\label{sec:Appendix3}
	
\begin{figure}[h]
		\centering
		\begin{subfigure}{\textwidth}
			\includegraphics[width=\textwidth]{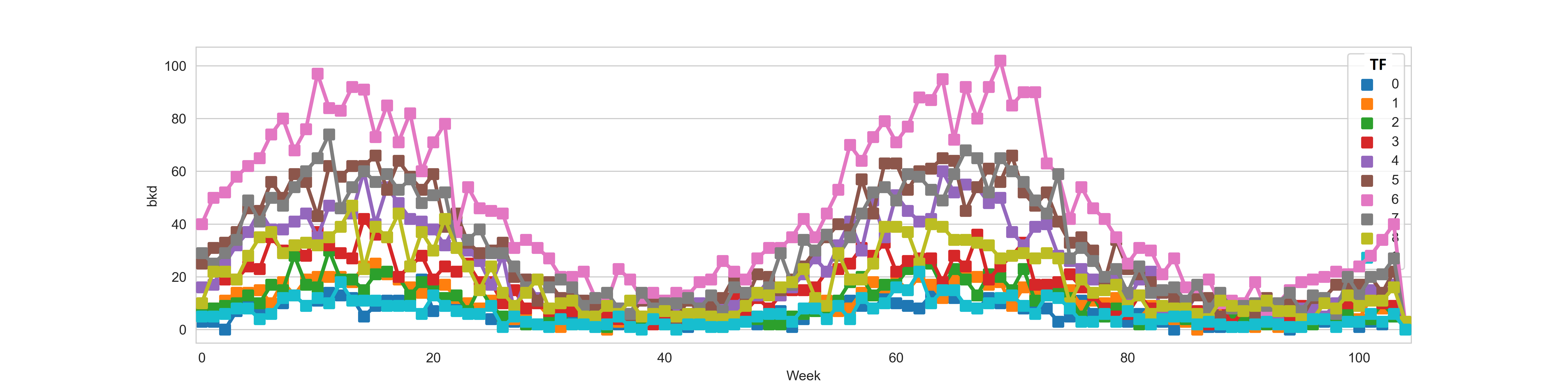}
			\caption{POS 0}
			\label{fig:dem_pos_0}
		\end{subfigure}
		\hfill
		\begin{subfigure}{\textwidth}
			\includegraphics[width=\textwidth]{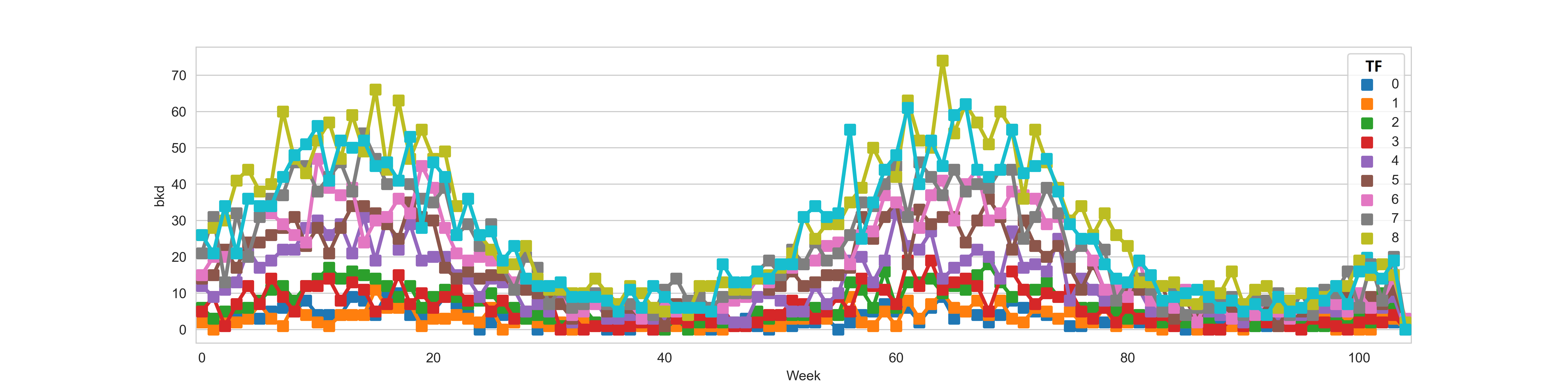}
			\caption{POS 1}
			\label{fig:dem_pos_1}
		\end{subfigure}
		\caption{Booking trends in simulated historical data}
		\label{fig:dem_plot}
	\end{figure}

	Figure \ref{fig:dem_plot} shows the trend of bookings over the two year simulated history, where we can observe the yearly seasonality marked by a peak and low season. Moreover, for POS 0, we observe more bookings in the middle TFs while for POS 1 we observe higher share of bookings for the later TFs. For both the POSs, the early TFs receive a relatively lower number of bookings.
	
\begin{figure}[h!]
		\centering
		\begin{subfigure}{.5\textwidth}
			\centering
			\includegraphics[width=1\linewidth]{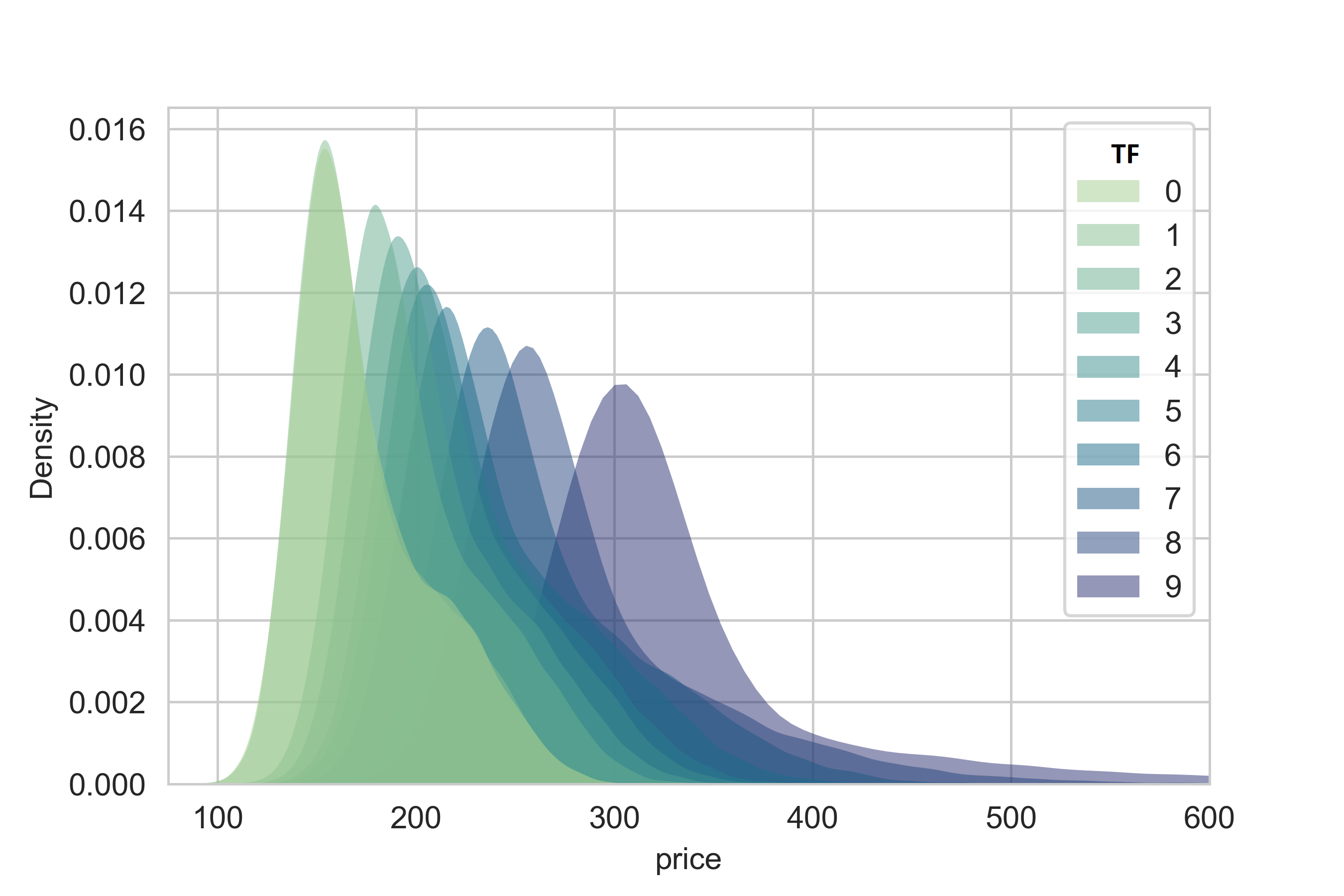}
			\caption{POS 0}
			\label{fig:prices_pos_0}
		\end{subfigure}%
		\begin{subfigure}{.5\textwidth}
			\centering
			\includegraphics[width=1\linewidth]{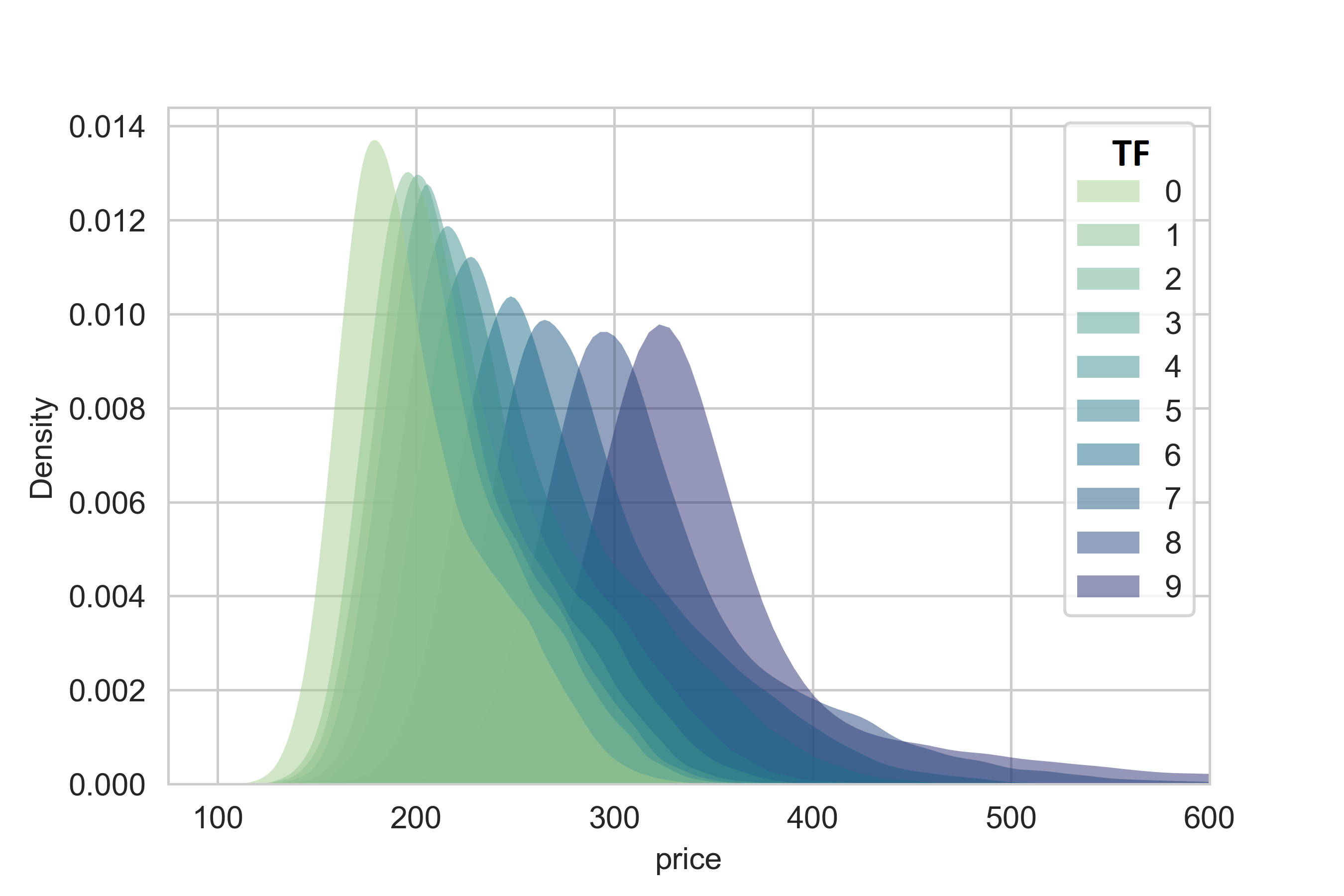}
			\caption{POS 1}
			\label{fig:prices_pos_1}
		\end{subfigure}
		\caption{Distribution of prices in simulated historical data}
		\label{fig:price_dist}
	\end{figure}

	Figure \ref{fig:price_dist} shows the distribution of prices over the two year history. We observe that since the willingness-to-pay ($\alpha$) is increasing as we get closer to the departure (higher TFs), the offered prices also show this increasing trend over TFs. Moreover, the average price for any TF is higher for POS 1 as compared to POS 0 again as the willingness-to-pay for POS 1 is higher than that for POS 0. 
	
	\clearpage

	\subsection{Additional results for the simulation in airline setting}\label{sec:Appendix4}
	
	\begin{table}[!ht]
		\centering
		\begin{tabular}{|ccc|cc|cc|}
			\hline
			POS & TF & $\alpha_{true} =   \frac{1}{\theta_{true}}$ & $\hat{\alpha}_{\tt{Two-stage-CO-Bayes}}   = \frac{1}{\hat{\mu}_{\theta}}$ & APE & $\hat{\alpha}_{\tt{Two-stage-CO-Freq}}   = \frac{1}{\hat{\theta}}$ & APE \\ \hline
			0 & 0 & 150 & 148.82 & 0.79\% & 150.74 & 0.49\% \\ \hline
			0 & 1 & 150 & 149.85 & 0.10\% & 155.31 & 3.54\% \\ \hline
			0 & 2 & 175 & 169.99 & 2.86\% & 176.81 & 1.03\% \\ \hline
			0 & 3 & 185 & 172.07 & 6.99\% & 178.27 & 3.64\% \\ \hline
			0 & 4 & 195 & 198.75 & 1.93\% & 206.34 & 5.81\% \\ \hline
			0 & 5 & 200 & 190.38 & 4.81\% & 197.21 & 1.40\% \\ \hline
			0 & 6 & 210 & 210.28 & 0.13\% & 218.04 & 3.83\% \\ \hline
			0 & 7 & 230 & 227.54 & 1.07\% & 236.44 & 2.80\% \\ \hline
			0 & 8 & 250 & 231.98 & 7.21\% & 240.42 & 3.83\% \\ \hline
			0 & 9 & 300 & 280.52 & 6.49\% & 290.69 & 3.10\% \\ \hline
			1 & 0 & 175 & 164.18 & 6.18\% & 166.00 & 5.14\% \\ \hline
			1 & 1 & 190 & 165.44 & 12.93\% & 171.56 & 9.71\% \\ \hline
			1 & 2 & 195 & 190.34 & 2.39\% & 198.18 & 1.63\% \\ \hline
			1 & 3 & 200 & 192.94 & 3.53\% & 200.02 & 0.01\% \\ \hline
			1 & 4 & 210 & 227.14 & 8.16\% & 236.04 & 12.40\% \\ \hline
			1 & 5 & 220 & 216.28 & 1.69\% & 224.16 & 1.89\% \\ \hline
			1 & 6 & 240 & 242.32 & 0.97\% & 251.48 & 4.78\% \\ \hline
			1 & 7 & 260 & 265.54 & 2.13\% & 276.27 & 6.26\% \\ \hline
			1 & 8 & 290 & 271.60 & 6.35\% & 281.72 & 2.85\% \\ \hline
			1 & 9 & 320 & 340.60 & 6.44\% & 353.32 & 10.41\% \\ \hline
			&  & \textbf{MAPE} & \textbf{} & \textbf{4.16\%} & \textbf{} & \textbf{4.22\%} \\ \hline
			&  & \textbf{wMAPE} & \textbf{} & \textbf{0.17\%} & \textbf{} & \textbf{0.21\%} \\ \hline
		\end{tabular}
		\caption{Comparison of price sensitivity parameters estimated via \texttt{Two-stage-CO} approach with Bayesian and frequentist method. Similar MAPE and wMAPE errors show that the improved estimation performance of \texttt{Two-stage-CO} method is not due to Bayesian estimation.}
		\label{tab:alpha_estimates_bayes_freq}
	\end{table}
	
%
%




\end{document}